\documentclass{article}

\usepackage[final]{neurips_2025}

\usepackage[utf8]{inputenc} % allow utf-8 input
\usepackage[T1]{fontenc}    % use 8-bit T1 fonts
\usepackage{hyperref}       % hyperlinks
\usepackage{url}            % simple URL typesetting
\usepackage{booktabs}       % professional-quality tables
\usepackage{amsfonts}       % blackboard math symbols
\usepackage{nicefrac}       % compact symbols for 1/2, etc.
\usepackage{microtype}      % microtypography
\usepackage{xcolor}         % colors
\usepackage{makecell}
\usepackage{pythonhighlight}
\usepackage{listings}

\usepackage{amssymb}
\usepackage{graphicx}
\usepackage{amsmath}
\usepackage{amsthm}
\usepackage{amsfonts}
\usepackage{algorithm}
\usepackage{algorithmicx}
\usepackage{algpseudocode}
\usepackage{subfigure}
\usepackage{wrapfig}
\usepackage{multirow}
\usepackage{makecell}
\usepackage{colortbl}

\usepackage{enumerate}
\usepackage{enumitem}

\title{GenPO: Generative Diffusion Models Meet \\On-Policy Reinforcement Learning}

\author{%
  Shutong Ding$^{1,5, *}$ \hspace{1em} Ke Hu$^{1,}$\thanks{Equal contribution. $\dag$Corresponding author.} \hspace{1em} Shan Zhong$^2$ \hspace{1em}  Haoyang Luo$^{1}$ \hspace{1em} Weinan Zhang$^3$ \\ \hspace{1em} \textbf{Jingya Wang$^{1,5}$ \hspace{1em} Jun Wang$^{4}$ \hspace{1em} Ye Shi$^{1,5,\dag}$ } \\ 
  \vspace{1pt}\\
  $^1$ShanghaiTech University \hspace{1em}
  $^2$University of Electronic Science and Technology of China\\
  $^3$Shanghai Jiao Tong University \hspace{1em}
  $^4$University College London\\
  $^5$MoE Key Laboratory of Intelligent Perception and Human Machine Collaboration\\
  \vspace{1pt}\\
  \texttt{ \{dingsht, huke2024, luohy12024\}@shanghaitech.edu.cn} \\
  \texttt{202211040927@std.uestc.edu.cn} \\
  \texttt{wnzhang@sjtu.edu.cn} \\
  \texttt{jun.wang@cs.ucl.ac.uk} \\
  \texttt{\{wangjingya, shiye\}@shanghaitech.edu.cn} \\
}

\begin{document}

\maketitle

\begin{abstract}
Recent advances in reinforcement learning (RL) have demonstrated the powerful exploration capabilities and multimodality of generative diffusion-based policies. While substantial progress has been made in offline RL and off-policy RL settings, integrating diffusion policies into on-policy frameworks like PPO remains underexplored. This gap is particularly significant given the widespread use of large-scale parallel GPU-accelerated simulators, such as IsaacLab, which are optimized for on-policy RL algorithms and enable rapid training of complex robotic tasks. A key challenge lies in computing state-action log-likelihoods under diffusion policies, which is straightforward for Gaussian policies but intractable for flow-based models due to irreversible forward-reverse processes and discretization errors (e.g., Euler-Maruyama approximations). To bridge this gap, we propose GenPO, a generative policy optimization framework that leverages exact diffusion inversion to construct invertible action mappings. GenPO introduces a novel doubled dummy action mechanism that enables invertibility via alternating updates, resolving log-likelihood computation barriers. Furthermore, we also use the action log-likelihood for unbiased entropy and KL divergence estimation, enabling KL-adaptive learning rates and entropy regularization in on-policy updates. Extensive experiments on eight IsaacLab benchmarks, including legged locomotion (Ant, Humanoid, Anymal-D, Unitree H1, Go2), dexterous manipulation (Shadow Hand), aerial control (Quadcopter), and robotic arm tasks (Franka), demonstrate GenPO's superiority over existing RL baselines. Notably, GenPO is the first method to successfully integrate diffusion policies into on-policy RL, unlocking their potential for large-scale parallelized training and real-world robotic deployment. The official implementation of GenPO is provided in \url{https://github.com/wadx2019/genpo/}.
\end{abstract}
\vspace{-1mm}
\section{Introduction}
\label{sect:intro}
\vspace{-3mm}
In recent years, generative diffusion models, such as DDPM~\cite{ho2020denoising}, DDIM~\cite{song2020denoising}, and flow matching~\cite{liu2022flow, lipmanflow}, have attracted considerable attention from reinforcement learning researchers due to their powerful exploration capabilities and multimodality compared with traditional parameterized Gaussian policies. However, existing works in this field almost merely focus on offline RL~\cite{wang2022diffusion, Ada_2024, kang2024efficient, chen2022offline,chen2023boosting,hansen2023idql, zhu2023madiff,ni2023metadiffuser} and off-policy RL~\cite{yang2023policy, ding2024diffusion, psenka2023learning, park2025flow, wang2024diffusion} methods, and the integration of diffusion or flow-based policies into on-policy reinforcement learning like PPO~\cite{schulman2017proximal} remains largely unexplored. However, existing massively parallel GPU-accelerated simulators, such as Isaac Gym~\cite{makoviychuk2021isaac} and its successor, IsaacLab, mainly~\cite{li2023parallel} benefit the training of on-policy reinforcement learning algorithms, especially PPO, which are capable of achieving substantial performance in a relatively short wall-clock time. This has resulted in a fundamental dilemma, hindering the practical deployment of existing diffusion/flow-based RL algorithms in real-world applications~\cite{chi2023diffusion, li2023crossway, pearce2023imitating, reuss2023goal, ding2025diopt}. 

Different from diffusion models~\cite{ding2024diffusion} applied in off-policy RL, the primary challenge in integrating flow policies with on-policy reinforcement learning is how to compute the log-likelihood of state-action pairs, while it is straightforward for Gaussian policies. The underlying reason here lies in the inconsistency and non-reversibility between the forward and reverse processes of the generative diffusion policy, caused by the Euler-Maruyama (EM) discretization~\cite{sarkka2019applied, celik2025dime}. Nevertheless, the optimization objective in on-policy RL necessitates the probability density of the given state-action pair. In this context, estimating the probability density of a given state-action pair in flow or diffusion models is nontrivial, and further utilizing this estimate for diffusion or flow policy improvement in the on-policy RL paradigm presents additional challenges.

To address these challenges, we propose GenPO, motivated by the exact diffusion inversion~\cite{wallace2023edict}, which constructs an invertible flow mapping and thus avoids the mismatch of forward and reverse processes of the diffusion policy. Specifically, we first construct a new Markov process with the action space twice the size of the original in GenPO due to the limitation of exact diffusion inversion, and then enable reversible flow processes by alternately updating the two parts of the dummy action. In that case, we can calculate the exact probability density of the given action in this new Markov decision problem via change of variables theory like normalizing flow~\cite{rezende2015variational, dinh2016density}. Moreover, the two parts of the dummy action will be averaged as one and then mapped to the original action space during inference. Hence, we can perform GenPO to update the generative diffusion policy in the reformulated Markov decision problem, but finally obtain the optimal solution to the original task. 
\begin{figure}
    \centering
    \vspace{-2mm}
    \includegraphics[width=1.0\linewidth]{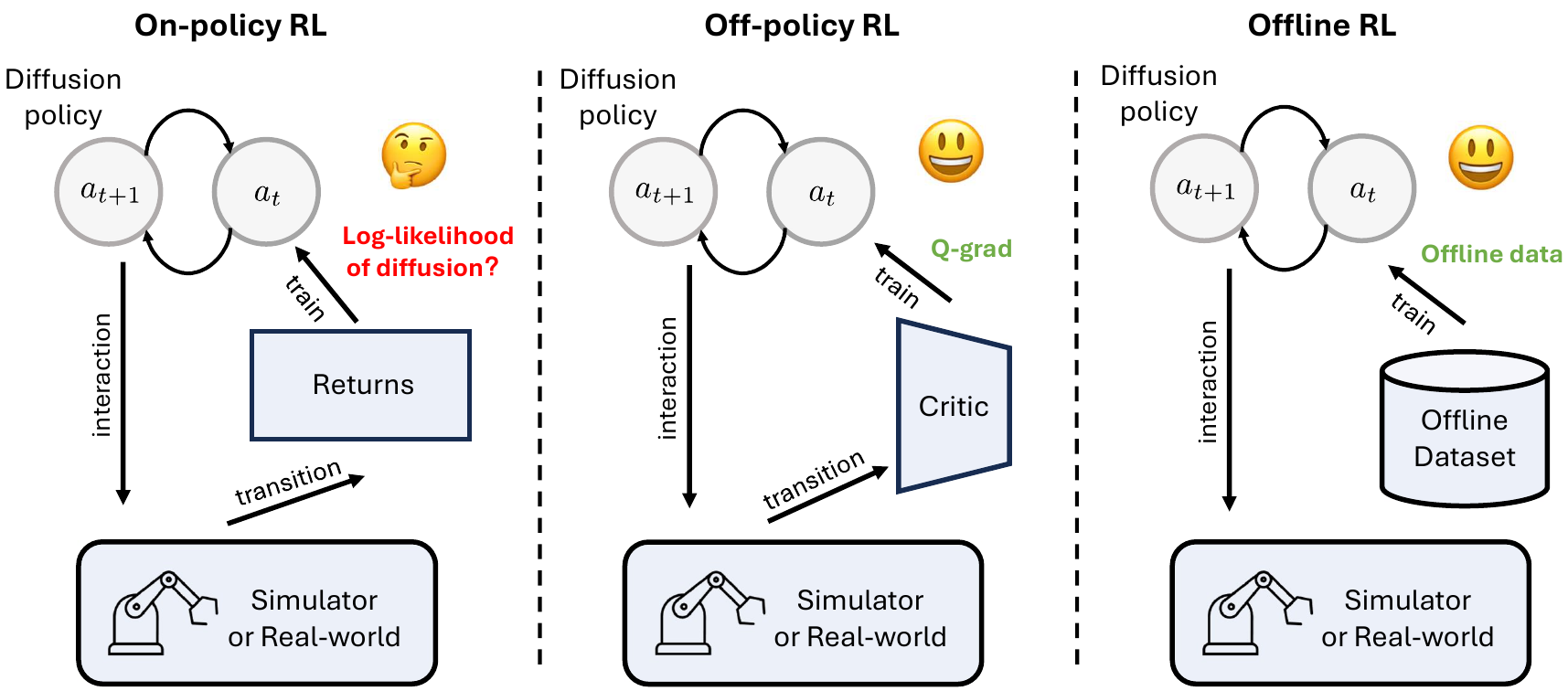}
    \caption{Existing diffusion-based reinforcement learning algorithms mainly focus on the off-policy (middle) and offline (right) RL. This is because we can generally obtain the gradient of the Q function to update the diffusion policy in off-policy RL and utilize the offline data to train the agent in offline RL. However, as for diffusion-based RL in the on-policy (left) algorithm, there still exists a challenge that we cannot obtain the log-likelihood of diffusion.}
    \vspace{-4mm}
    \label{fig:motivation}
\end{figure}

In addition, with the exact probability density, we estimate~\cite{cover1999elements} the entropy of the diffusion policy and the Kullback-Leibler (KL) divergence with respect to the behavior diffusion policy. This allows for the combination of entropy regularization and KL-adaptive learning rate adjustment into GenPO, which has been confirmed effective in PPO. To demonstrate the superiority of GenPO, we conduct experiments on 8 IsaacLab benchmarks~\cite{makoviychuk2021isaac}, which cover robot ant, humanoid, quadcopter, Franka robot arm~\cite{haddadin2024franka}, Shadow dexterous hand~\cite{tuffield2003shadow}, ANYbotics anymal-D, and Unitree legged robots~\cite{zhuang2024humanoid}. 
The final results show that the proposed GenPO outperforms previous RL baselines in cumulative returns, while maintaining comparable sample efficiency and faster convergence. It is worth noting that GenPO is the first trial to incorporate the generative diffusion policy into on-policy RL algorithms and opens brand-new avenues for diffusion-type policies applied in large-scale parallel simulators and real-world robot control. Our contributions are summarized as follows:
\begin{itemize}[leftmargin=4pt, rightmargin=4pt]
\item \textbf{Bridging generative diffusion models and on-policy RL.} GenPO achieves the first stable integration of diffusion-based policies with on-policy reinforcement learning, successfully deploying the inherent exploration capabilities and multimodality of diffusion models in large-scale GPU parallel simulators (IsaacLab). 
\item \textbf{Closing the likelihood gap in diffusion policies.} Unlike Gaussian policies with closed-form densities, diffusion policies lack tractable likelihoods. GenPO overcomes this by using invertible diffusion dynamics, enabling 1) exact log-likelihood computation, 2) unbiased entropy estimation, and 3) analytical KL divergence-bringing Gaussian-style advantages to expressive diffusion models.
\item \textbf{Comprehensive experiments on IsaacLab benchmarks.} We evaluated GenPO on IsaacLab benchmarks. The experimental results show that in a massively parallel environment, previous diffusion-based reinforcement learning algorithms are almost ineffective, while GenPO achieves the best performance in terms of sample utilization efficiency and episodic rewards, far surpassing other algorithms.
\end{itemize}

% \begin{itemize}[leftmargin=4pt, rightmargin=4pt]
%     \item \textbf{Bridging generative diffusion models and on-policy RL.} GenPO constitutes the first integration of diffusion-based policies into on-policy reinforcement learning, facilitating powerful exploration capability and multimodality of diffusion models in large-scale simulators like IsaacLab. 
%     \item \textbf{Calculating log-likelihood with exact diffusion inversion.} Motivated by exact diffusion inversion, GenPO enables tractable computation of the log-likelihood for a given action within a reformulated Markov decision process, which is non-trivial for generative diffusion-based policies. 
%     \item \textbf{Unbiased estimation of entropy and KL divergence.} We also develop unbiased estimation techniques for the entropy and the KL divergence flow policies to supplement GenPO to achieve better performance, which are inaccessible in previous diffusion-based RL methods.
%     \item \textbf{Comprehensive experiments on IsaacLab benchmarks.} We demonstrate the efficiency of GenPO with comprehensive experiments on 8 IsaacLab benchmarks, including but not limited to robot ant, robot arm, dexterous hand, humanoid, quadrotor, and quadcopter.
% \end{itemize}

\section{Related Works}

\label{sect:related works}
In this section, we review the literature of generative models such as VAE~\cite{kingma2013auto}, GAN~\cite{creswell2018generative}, normalizing flow~\cite{rezende2015variational}, diffusion~\cite{zhang2021diffusion}, and flow-based model~\cite{liu2022flow, lipman2022flow} for policy learning, and generally divide them into two classes according to their learning paradigm.

\textbf{Generative Policy for Online Reinforcement Learning.} The goal of online RL~\cite{sutton2018reinforcement} is to learn an optimal policy with the interactions of the given environment. It is challenging to train a generative model policy within an online RL paradigm due to the absence of the action label. Generative policy learning methods in online RL can be broadly grouped into two paradigms. The first paradigm views the generative policies as a black-box function approximation like neural networks, and optimizes them via the policy gradient. In this paradigm, representative methods include DACER~\cite{wang2024diffusion}, CPQL~\cite{chen2024boosting}, FlowPG~\cite{brahmanage2023flowpg}, SAC-NF~\cite{mazoure2020leveraging}, and TRPO-NF~\cite{tang2018boosting}, which directly apply deterministic or policy gradient on the generative model policies. By contrast, the second category leverages the internal structure and mechanisms of the generative model itself to perform the policy improvement. For instance, DIPO~\cite{yang2023policy} and QVPO~\cite{ding2024diffusion} respectively utilize the gradient and the value magnitude of the Q-function to identify optimal actions, which are then applied to the variational loss of the diffusion model. Additionally, QSM~\cite{psenka2023learning} and MaxEntDP~\cite{dong2025maximum} regard the noise network in the diffusion model as the score function of the target policy distribution and then employ the gradient of the Q-function and its variants to train the noise network. As to normalizing flow-based approaches, MEow~\cite{chao2024maximum} exploits the layer-wise nonlinear property of the flow network to train the Q and V functions, which yields the corresponding maximum-entropy normalizing flow policy based on SAC~\cite{haarnoja2018soft, haarnoja2018soft2}. Overall, current approaches that employ generative-model policies in online RL have predominantly focused on integrating these models with off-policy RL algorithms like SAC, with few efforts to combine them with on-policy RL methods.

\textbf{Generative Policy for Other Learning Paradigms.} Compared with generative policy for online RL, more existing research works on generative policy have been devoted to offline RL~\cite{levine2020offline} and imitation learning paradigms, wherein policies are learned from offline datasets without environment interaction. For offline RL, BCQ~\cite{fujimoto2019off}, PLAS~\cite{zhou2021plas}, SPOT~\cite{wu2022supported} adopt VAE~\cite{kingma2013auto} policy regularizer to prevent the policy from optimizing outside the support of the dataset. Besides, CPED~\cite{zhang2023constrained} utilizes flow GAN~\cite{grover2017flow} to achieve more accurate policy regularization. Furthermore, diffusion-QL~\cite{wang2022diffusion}, EDP~\cite{kang2024efficient}, SRDP~\cite{Ada_2024}, CEP~\cite{lu2023contrastive}, and FQL~\cite{park2025flow} leverage diffusion~\cite{ho2020denoising} and flow~\cite{liu2022flow, lipman2022flow} policies to accurately model the policy distribution in offline datasets, thereby obtaining policies with superior generalization capability. Regarding imitation learning, diffusion policy~\cite{chi2023diffusion} and diffuser~\cite{janner2022planning} serve as canonical examples of using diffusion models to fit the state-action mapping and state-trajectory mapping, respectively. The latter one is also referred to as the diffusion planner. Subsequently, FMIL~\cite{rouxel2024flow}, AVDC~\cite{ko2023learning}, and decision diffuser~\cite{ajay2022conditional} follow their steps and demonstrate enhanced performance in different robots. Furthermore, DPPO~\cite{ren2024diffusion} applies policy gradient to fine-tune the diffusion policy trained with the offline dataset. Nonetheless, DPPO merely explores the use of the RL technique in finetuning an offline-pretrained diffusion policy, rather than representing a diffusion-based online RL algorithm. 

However, all of the above generative policy learning approaches have not been effectively integrated with on-policy reinforcement learning algorithms, making them unable to sufficiently leverage modern massively parallel GPU-accelerated simulators. Specifically, off-policy algorithms often struggle with convergence in such simulators, while offline RL and imitation learning methods are fundamentally incompatible with them due to the lack of online interaction. In contrast, our proposed Generative diffusion Policy Optimization (GenPO) seamlessly integrates with on-policy algorithms, enabling efficient training of high-performing generative policies within a short wall-clock time using large-scale GPU-parallelized simulation platforms.

\section{Preliminaries}
\label{sect:preliminaries}
\vspace{-2mm}
\subsection{On-policy Reinforcement Learning}
\label{sect:ppo}

Reinforcement learning problems~\cite{sutton2018reinforcement} are formalized as Markov decision processes (MDPs), defined by the tuple $(\mathcal{S}, \mathcal{A}, p, r, \rho_0, \gamma)$, where $\mathcal{S}$ is the state space, $\mathcal{A}$ is the action space, $p(s' \mid s,a): \mathcal{S}\times \mathcal{S}\times \mathcal{A}\rightarrow [0, \infty)$ denotes the transition dynamics, $r(s, a): \mathcal{S}\times \mathcal{A}\rightarrow \mathbb{R}$ is the reward function, $\rho_0(s): \mathcal{S} \rightarrow [0, \infty)$ is the distribution of the initial state, and $\gamma\in[0,1)$ is the discount factor for the value estimation. The goal of RL is to maximize the expected discounted return as $J(\theta) \;=\; \mathbb{E}_{\tau\sim\pi_{\theta}}\left[\sum_{t=0}^{\infty}\gamma^t r_t\right]$, where $\pi_{\theta}(a\!\mid\!s)$ is the behavior of the agent parametrized by $\theta$, and induces a corresponding distribution over behavior trajectories $\tau$. For convenience, RL also defines two different value functions: the state value function $V_\pi({s}) = \mathbb{E}_{\tau\sim\pi}[\sum_{t=0}\gamma^tr_{t}|{s}_0={s}]$ and the state-action value function $Q_\pi(s, a) = \mathbb{E}_{\tau\sim\pi}[\sum_{t=0}\gamma^tr_{t}|{a}_0={a},{s}_0={s}]$. Besides, the advantage function, which represents the benefit of taking one specific action compared with the expected return at the given state, is defined as $A_\pi({s},{a}) = Q_\pi({s}, {a}) - V_\pi({s})$.

On-policy RL algorithms train the policy with the transitions generated by the current policy itself, while off-policy RL algorithms use the transitions from a different policy as training samples. While on-policy RL algorithms preserve the consistency between the behavior and target policies, but often suffer from high variance and sample inefficiency. Proximal Policy Optimization (PPO)~\cite{schulman2017proximal} addresses these challenges by introducing the clipped surrogate objective:
\begin{equation}
\mathcal{L}^{\mathrm{CLIP}}(\theta)
:=\mathbb{E}_{(s_t,a_t)\sim\pi_{\theta_{\mathrm{old}}}}
\left[\min\bigl(r_t(\theta)\hat{A}_t,\;\mathrm{clip}(r_t(\theta),1-\epsilon,1+\epsilon)\hat{A}_t\bigr)\right],
\end{equation}
where $r_t(\theta)=\frac{\pi_\theta(a_t\mid s_t)}{\pi_{\theta_{\mathrm{old}}}(a_t\mid s_t)}$ and $\hat{A}_t$ is the estimation of the advantage function via GAE~\cite{schulman2015high}. By leveraging the importance sampling technique and trust-region insights~\cite{schulman2015trust} without second-order complexity for approximation of the KL divergence constraint, PPO can achieve stable and efficient policy updates. The simplicity, robustness, and strong empirical performance on continuous-control benchmarks~\cite{makoviychuk2021isaac, zhuang2024humanoid} of PPO have made it the first algorithm to be adopted in many domains.

\subsection{Generative Diffusion Models and Flow Matching}
\label{sect:diffusion}
Generative diffusion models~\cite{ho2020denoising, song2020score, song2020denoising} are a kind of powerful latent variable generative model that can transform samples from the standard Gaussian distribution $x_{T}\sim \mathcal{N}(0, I)$ to the data distribution $x_0\sim p_{data}$ via the denoising procedure with the noise network. The relationship between $x_0$ and $x_T$ is defined by the forward SDE of diffusion 
\begin{equation}
\mathrm{d}x_t = f(x_t,t)\,\mathrm{d}t + g(t)\,\mathrm{d}W_t,
\end{equation}
where $W_t$ the standard Wiener process (a.k.a., Brownian motion), $f(x_t,t)$ is a vector-valued function called the drift coefficient of $x_t$, and $g(t)$is a scalar function known as the diffusion coefficient of $x_t$. According to \cite{song2020score}, the reverse-time SDE of diffusion (i.e., denoising procedure) is
\begin{equation}
\mathrm{d}x_t = \bigl[f(x_t,t) - g(t)^2\nabla_x\log p_t(x_t)\bigr]\,\mathrm{d}t + g(t)\,\mathrm{d}\overline{W}_t,
\end{equation}
where $\overline{W}_t$ is a standard Wiener process when time flows backwards from $T$ to $0$. In practice, DDPM~\cite{ho2020denoising} discretizes the SDE and approximates the forward step as $q(x_t \mid x_{t-1}) := \mathcal{N}(x_t; \sqrt{1-\beta_t} x_{t-1}, \beta_tI)$, where $\beta_t$ is the variance schedule. The noise network $\boldsymbol{\epsilon}_\theta(x_t, t)$ in DDPM can be trained by the variational lower bound loss:
\begin{equation}
\mathbb{E}_{t \sim [1,T], x_0, \boldsymbol{\epsilon}_t}\left[||\boldsymbol{\epsilon}_t - \boldsymbol{\epsilon}_\theta(\sqrt{\bar{\alpha}_t} x_0 + \sqrt{1 - \bar{\alpha}_t} \boldsymbol{\epsilon}_t, t)||^2 \right],
\end{equation}
where $\alpha_t = 1 - \beta_t$, $\bar{\alpha}_t = \prod_{s=0}^t \alpha_s$, and $\boldsymbol{\epsilon}_t\sim \mathcal{N}(0, I)$.

The flow matching model~\cite{liu2022flow,lipman2022flow} is developed for faster generation compared with classical diffusion models like DDPM, and can be viewed as a special type of generative diffusion model~\cite {gaodiffusion}.
Similar to DDPM, flow matching model learns a time-dependent vector field $v_\theta(x,t)$, i.e., noise network in DDPM, that transports samples along a predefined family of conditional distributions $\{p_t\}_{t\in[0,1]}$ satisfying $x_1 \sim p_{\mathrm{data}}$ and $x_0 \sim \mathcal{N}(0, I)$. Hence, the sample from the target distribution $p_{data}$ can be achieved via solving the ode $x_1 = x_0 + \int_{0}^{1}v(x_t,t)dt$.

Besides, The loss of flow matching is given as $\mathcal{L}(\theta)=\int_0^1 \mathbb{E}_{x\sim p_t}\left\|v_\theta(x,t)-v^*(x,t)\right\|^2\,\mathrm{d}t$, where $v^*(x_t,t)$ is the true velocity field under the chosen path. When the path is Gaussian diffusion, $v^*(x,t)$ coincides with the score-drift term of the reverse SDE in diffusion. However, the log-likelihood of the given sample cannot be calculated directly in the generative diffusion framework.

\section{GenPO: An On-policy RL Method based on Diffusion Models}
\label{sect:method}
\begin{figure}
    \centering
    \vspace{-3mm}
    \includegraphics[width=0.9\linewidth]{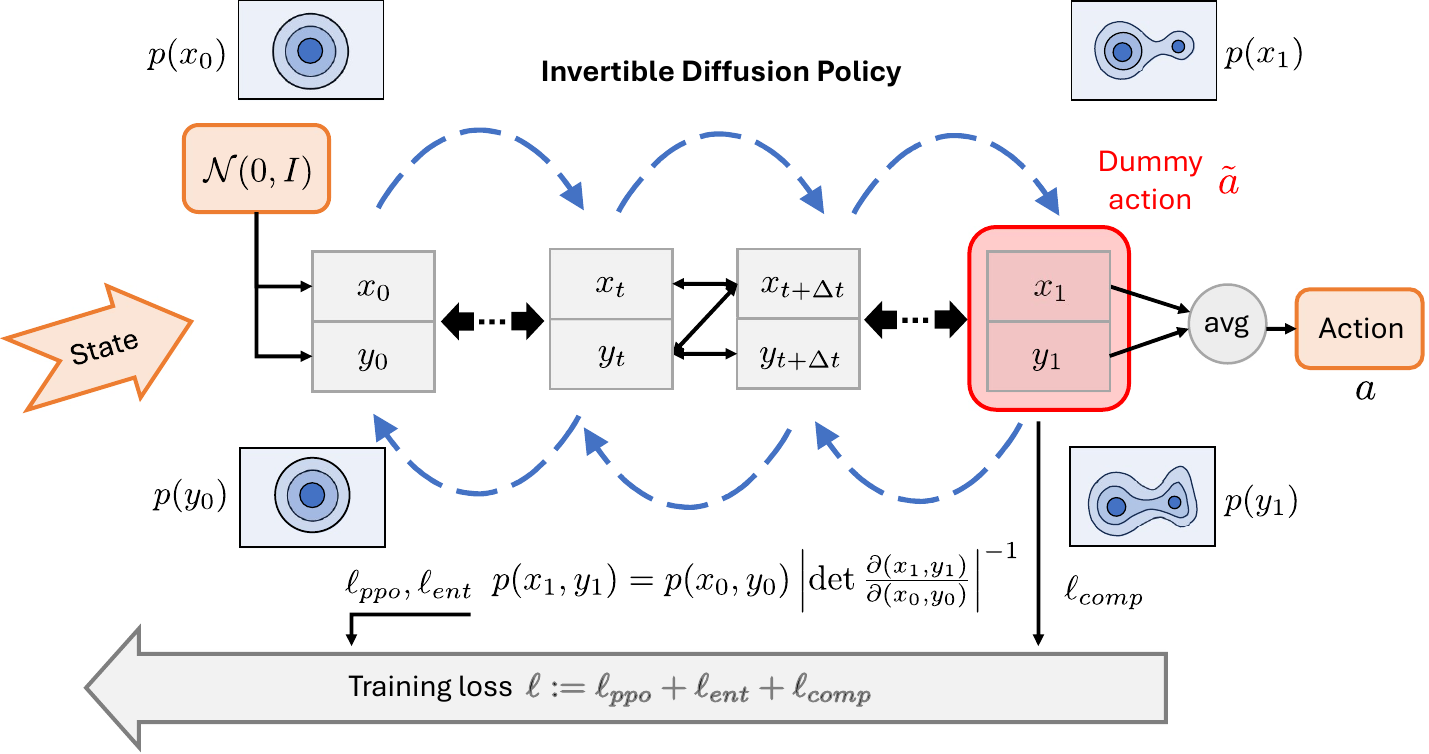}
    \vspace{-3mm}
    \caption{Forward and reverse process of GenPO. The forward process is to sample actions with the given state; the reverse process is to compute the probability density of the given state-action pair. Notably, the forward and reverse processes are invertible.}
    \label{fig:fpo}
    \vspace{-3mm}
\end{figure}
As mentioned in Section~\ref{sect:diffusion}, the log-likelihood of the given samples cannot be computed directly in generative models. However, to integrate diffusion policy into on-policy RL, we must present the explicit formula of log-likelihood under the given state-action pair for the policy update. Hence, we first elucidate the intrinsic reasons that make the computation of the log-likelihood in diffusion models challenging. Then, we address these challenges by (1) applying the exact diffusion inversion technique~\cite{wallace2023edict} and utilizing the change of variables technique like normalizing flow~\cite{rezende2015variational}, (2) constructing a reformulated Markov decision process with a modified action space, along with a trick that maps the generated actions back to the original action space.

Besides, based on this framework, we also introduce an unbiased estimation method for both the entropy of the diffusion policy and the Kullback-Leibler (KL) divergence between different generative diffusion policies. Notably, this has not yet been accomplished by previous diffusion-based RL methods. In addition, since policy update is actually performed on a reformulated MDP, extraneous exploration is induced, thereby degrading RL learning efficiency. To address this issue, we also introduce an auxiliary compression loss term. Leveraging these techniques, we finally present the practical implementation of the proposed GenPO algorithm, which successfully incorporates the generative policy to PPO~\cite{schulman2017proximal} in an effective manner. Figure~\ref{fig:fpo} shows the training and inference process of the proposed GenPO.

\subsection{Challenges in Calculating Diffusion Probability}
\label{4.1}
\vspace{-2mm}
Recalling existing generative models, such as VAE~\cite{kingma2013auto}, GAN~\cite{creswell2018generative}, normalizing flow~\cite{rezende2015variational}, and diffusion model~\cite{song2020score}, it can be found that only normalizing flow and its variants like flow GAN~\cite{grover2017flow} allow the exact likelihood computation via the change of variables theorem~\cite{rezende2015variational}. In contrast, for other generative models, probability densities or related statistics can only be obtained approximately via special designs~\cite{ding2024diffusion, celik2025dime, wang2024diffusion}. Applying such approximate probability density estimates for policy update is unacceptable in on-policy RL algorithms.

According to Lemma~1, if we want to compute the probability density of a generative model exactly, the generative model must be invertible between the sampling distribution, like the standard Gaussian distribution, and the target distribution. Thus, we consider designing a reversible generative diffusion model and employing the change of variables theorem (\ref{eq:change}) to compute its probability density.

\textbf{Lemma 1.}
% \label{lemma 1}
\textit{\textbf{(Change of Variables ~\cite{blitzstein2019introduction})} Let $f: \mathbb{R}^n\rightarrow \mathbb{R}^n$ is an invertible and smooth mapping. If we have the random variable $X\sim q(x)$ and the random variable $Y=f(X)$ transformed by function $f$, the distribution $p(y)$ of $Y$ is
\begin{equation}
    p(y)=q(x)\left|\det\frac{\partial f}{\partial x}\right|^{-1}
    \label{eq:change}
\end{equation}
}

However, as mentioned in \cite{wallace2023edict, wang2024belm, zhang2024exact}, it is nontrivial to realize an exact invertible diffusion model. This is because of the inconsistency between the forward and reverse processes of the generative diffusion model. (\ref{eq:ddim}) shows the forward and reverse process of DDIM~\cite{song2020denoising}. It can be observed that this inconsistency is caused by the approximation of $x_t$ with $x_{t-1}$ in the forward process, since the $x_t$ is unavailable in the forward step $t-1$.
\begin{equation}
    \begin{aligned}
            \text{Reverse:} \quad& x_{t-1} = \sqrt{\alpha_{t-1}}\frac{x_t-\sqrt{1-\alpha_t}\boldsymbol{\epsilon}_\theta(x_t,t)}{\sqrt{\alpha_t}}+\sqrt{1-\alpha_{t-1}}\boldsymbol{\epsilon}_\theta(x_t,t)\\
            \text{Forward:} \quad& x_t = \frac{x_{t-1}-b_t\boldsymbol{\epsilon}_\theta(x_t,t)}{a_t}\approx \frac{x_{t-1}-b_t\boldsymbol{\epsilon}_\theta(x_{t-1},t)}{a_t} \\
    \end{aligned}
    \label{eq:ddim}
\end{equation}
where $a_t=\sqrt{\alpha_{t-1}/\alpha_t}, b_t=-\sqrt{\alpha_{t-1}(1-\alpha_t)/\alpha_t}+\sqrt{1-\alpha_t}$.

\subsection{Exact Diffusion Inversion in Reformulated MDP}
\label{4.2}

Consequently, to realize an invertible diffusion model, this issue must be addressed. Motivated by EDICT~\cite{wallace2023edict}, we realize the invertible diffusion model via maintaining two coupled noise vectors and updating them alternately in the forward and reverse processes of the generative diffusion model, and then extend it into flow matching~\cite{lipmanflow, liu2022flow} for fast generation as shown in (\ref{eq:fpo_reverse}, \ref{eq:fpo_forward}). However, the two coupled noise vectors also lead to a doubled sample space. This implies we cannot directly apply this technique to diffusion policy when the dimension of the action space is odd. 

To resolve this problem, we reformulate the original MDP problem with a doubled dummy action space $\tilde{\mathcal{A}}$ in GenPO. Each dummy action $\tilde{a} = (x, y)$ consists of two components, which are subsequently averaged to produce a single action $a = \frac{x+y}{2}$ in the original space. It is obvious that, when the policy in the reformulated MDP is optimal, it is also optimal with the average mapping in the original MDP problem. 
\begin{equation}
    \begin{array}{lll}
            \text{Reverse:} \quad& \tilde{x}_{t+\Delta t} = x_t + v_\theta(y_t,t)\Delta t, & \tilde{y}_{t+\Delta t} = y_t + v_\theta(\tilde{x}_{t+\Delta t},t)\Delta t \\
            \text{Mixing:} \quad& x_{t+\Delta t} = p\cdot \tilde{x}_{t+\Delta t} + (1-p)\cdot \tilde{y}_{t+\Delta t}, & y_{t+\Delta t}= p \cdot \tilde{y}_{t+\Delta t} + (1-p) \cdot x_{t+\Delta t} \\
    \end{array}
    \label{eq:fpo_reverse}
\end{equation}

Notably, different from EDICT, which just applies exact diffusion inversion for the inference of diffusion, GenPO independently samples the standard Gaussian noise for $x_0, y_0$ rather than samples one noise $\epsilon$ and sets $x_0=y_0=\epsilon$ to allow for the probability calculation using changes of variables (\ref{eq:change}). Moreover, due to the doubled action space in the modified MDP, there exists a meaningless exploration problem in the diffusion policy. For instance, if the optimal action $a^\star=0$, we have $\tilde{a}=(-1,1),(-2,2)$ are both the optimal dummy action. Hence, to prevent unnecessary exploration in the modified MDP, a mixing scheme (\ref{eq:fpo_reverse}, \ref{eq:fpo_forward}) is adopted so that the two parts $x,y$ of dummy action components remain closely aligned. In (\ref{eq:fpo_reverse}, \ref{eq:fpo_forward}), the mixing scheme is employed to facilitate information exchange between the $x$ and $y$ components of $\tilde{a}$, thereby ensuring the diffusion policy converges to dummy actions with minimal discrepancy between $x$ and $y$ parts. Here the coefficient $p$ is used to control the intensity of the interchanged information.
\begin{equation}
    \begin{array}{lll}
            \text{Unmixing:} \quad & \tilde{y}_{t+\Delta t} = \frac{y_{t+\Delta t}-(1-p)x_{t+\Delta t}}{p} & \tilde{x}_{t+\Delta t} = \frac{x_{t+\Delta t}-(1-p)\tilde{y}_{t+\Delta t}}{p}\\
            \text{Forward:} \quad& y_t = \tilde{y}_{t+\Delta t} - v_\theta(\tilde{x}_{t+\Delta t},t)\Delta t, & x_t = \tilde{x}_{t+\Delta t} - v_\theta(y_{t},t)\Delta t 
            
    \end{array}
    \label{eq:fpo_forward}
\end{equation}

\subsection{Practical Implementation}
\label{sect:practical}

To enhance GenPO's exploration and stabilize training, we estimate policy entropy and introduce an adaptive KL-divergence-based learning rate schedule for flow policies.
Unlike Gaussian policies, flow policies lack closed-form expressions for entropy and KL divergence, preventing direct computation. As a result, prior diffusion RL algorithms~\cite{ding2024diffusion, wang2024diffusion, celik2025dime} have relied on heuristic approximations to encourage exploration.
In contrast, GenPO is exactly invertible (Section~\ref{4.2}), allowing precise computation of the log-likelihood of any state-action pair via the change-of-variables (\ref{eq:change}) in Section~\ref{4.1}. This enables unbiased estimation of both entropy and KL divergence. Accordingly, we define the entropy loss in Eq.~\ref{eq:ent}.
\begin{equation}
    \mathcal{L}^{ENT}(\pi_\theta):=\mathbb{E}_{s,\tilde{a}\sim\pi_{\theta}}\left[\log\left(\pi_{\theta}(\tilde{a}\mid s)\right)\right].
    \label{eq:ent}
\end{equation}
Besides, to enable adaptive scheduling of the learning rate, we also present an unbiased estimation of the KL divergence as shown in Algorithm~\ref{alg:train}, and the estimation formula for computing the KL divergence between diffusion policies is
\begin{equation}
    \widehat{\text{KL}}(\pi_{\theta_{old}}\mid \pi_\theta):=\mathbb{E}_{s,\tilde{a}\sim\pi_{\theta_{old}}}\left[\log\left(\pi_{\theta_{old}}(\tilde{a}\mid s)\right)-\log\left(\pi_\theta(\tilde{a}\mid s)\right)\right].
    \label{eq:kl}
\end{equation}

We also present an instantiation of our algorithm within the PPO framework. The surrogate loss of PPO is modified as (\ref{eq:ppo}).
\begin{equation}
\mathcal{L}^{\mathrm{PPO}}(\theta)
:=\mathbb{E}_{(s_t,\tilde{a}_t)\sim\pi_{\theta_{\mathrm{old}}}}
\left[\min\left(\frac{\pi_{\theta}(\tilde{a}_t \mid s_t)}{\pi_{\theta_{old}}(\tilde{a}_t\mid s_t)}\hat{A}_t,\;\mathrm{clip}(\frac{\pi_{\theta}(\tilde{a}_t \mid s_t)}{\pi_{\theta_{old}}(\tilde{a}_t\mid s_t)},1-\epsilon,1+\epsilon)\hat{A}_t\right)\right].
\label{eq:ppo}
\end{equation}
Notably, to mitigate this issue and prevent ineffective exploration of the action space, we introduce a compression loss as $\mathbb{E}_{x_1,y_1\sim \pi_\theta}\left[\left(x_1-y_1\right)^2\right]$, which diminishes the mean square error between $x$ and $y$ components, in the policy loss. The final diffusion policy loss is shown in (\ref{eq:loss}).
\begin{equation}
\mathcal{L}(\theta):= \mathcal{L}^{PPO} + \lambda\mathcal{L}^{ENT} +  \nu \mathbb{E}_{x_1,y_1\sim \pi_\theta}\left[\left(x_1-y_1\right)^2\right],
    \label{eq:loss}
\end{equation}
where $\lambda$ and $\nu$ are the coefficients of the entropy and compression loss, respectively.
\begin{algorithm*}[ht!]
	\caption{Generative Diffusion Policy Optimization} 
	\label{alg:train} 
        \textbf{Input:} generative diffusion policy $\pi_\theta(\tilde{a}\mid s)$, value network $V_{\omega}(s)$.
	\begin{algorithmic}[1] 
            \State $\theta^\prime \gets \theta$
            \For{$t$ \textbf{in} $1, 2, \cdots, T$}
                \For{each actor}
                    \State Sample dummy actions $\tilde{a}=(x,y)$ from $\pi_{\theta}(a\mid s)$
                    \State Use $a=\frac{x+y}{2}$ to interact with the environment for $N$ timesteps
                    \State Calculate the probability of $\pi_{\theta}(\tilde{a}\mid s)$ of the executed dummy actions
                    \State Use GAE to estimate advantages $\hat{A}_1, \cdots, \hat{A}_N$ and returns $\hat{R}_1,\cdots, \hat{R}_n$
                    \State Store transitions, dummy actions, probability, and advantage value in rollout 
                \EndFor
                \For{$k$
                \textbf{in} $1, 2, \cdots, K$}
                \State Estimate KL divergence according to (\ref{eq:kl}) and adjust the learning rate $\alpha$
                \[
                \alpha \leftarrow \left\{\begin{array}{cc}
                   \frac{\alpha}{2},  &  D_{KL}\ge \overline{\epsilon}; \\
                    2\alpha, & D_{KL}\le \underline{\epsilon}.
                \end{array}\right.
                \]
                \State Update $\pi_\theta(\tilde{a}\mid s)$ with the surrogate loss (\ref{eq:loss}) with and mini-batches from rollout
                \State Update the value network $V_{\omega}(s)$ with loss $\mathbb{E}_{s,a\sim \pi_\theta}\left[\left\|V_{\omega}(s) - \hat{R}(s,a)\right\|^2\right]$
                \EndFor
                \EndFor
	\end{algorithmic} 
\end{algorithm*}

\section{Experiments}
\label{sect: experiment}
In this section, we empirically evaluate the proposed method on several control tasks within IsaacLab. We also conduct ablation studies to investigate the impact of key hyperparameters. These experiments aim to compare our approach with popular online RL algorithms and analyze how hyperparameter choices affect performance.
\subsection{Comparative Evaluation}

\textbf{Setup.} To evaluate our method, we conducted experiments on a suite of benchmark tasks provided by IsaacLab. The benchmark environments used in this study include Isaac-Ant-v0, Isaac-Humanoid-v0, Isaac-Lift-Cube-Franka-v0, Isaac-Repose-Cube-Shadow-Direct-v0, Isaac-Velocity-Flat-Anymal-D-v0, Isaac-Velocity-Rough-Unitree-Go2-v0, Isaac-Velocity-Rough-H1-v0, and Isaac-Quadcopter-Direct-v0. For brevity, we refer to these environments respectively as \textbf{Ant}, \textbf{Humanoid}, \textbf{Franka Arm}, \textbf{Shadow Hand}, \textbf{Anymal-D}, \textbf{Unitree-Go2}, \textbf{Unitree-H1}, and \textbf{Quadcopter} throughout the remainder of the paper.
Our algorithm GenPO is compared and evaluated against several well-known model-free algorithms, including DDPG, TD3, SAC, PPO, as well as two diffusion-based off-policy algorithms: DACER and QVPO.
To assess statistical robustness, all experiments are repeated across five random seeds.
More details related to the experiments can be found in the supplementary materials.
\begin{figure}
    \centering
    \includegraphics[width=1.0\linewidth]{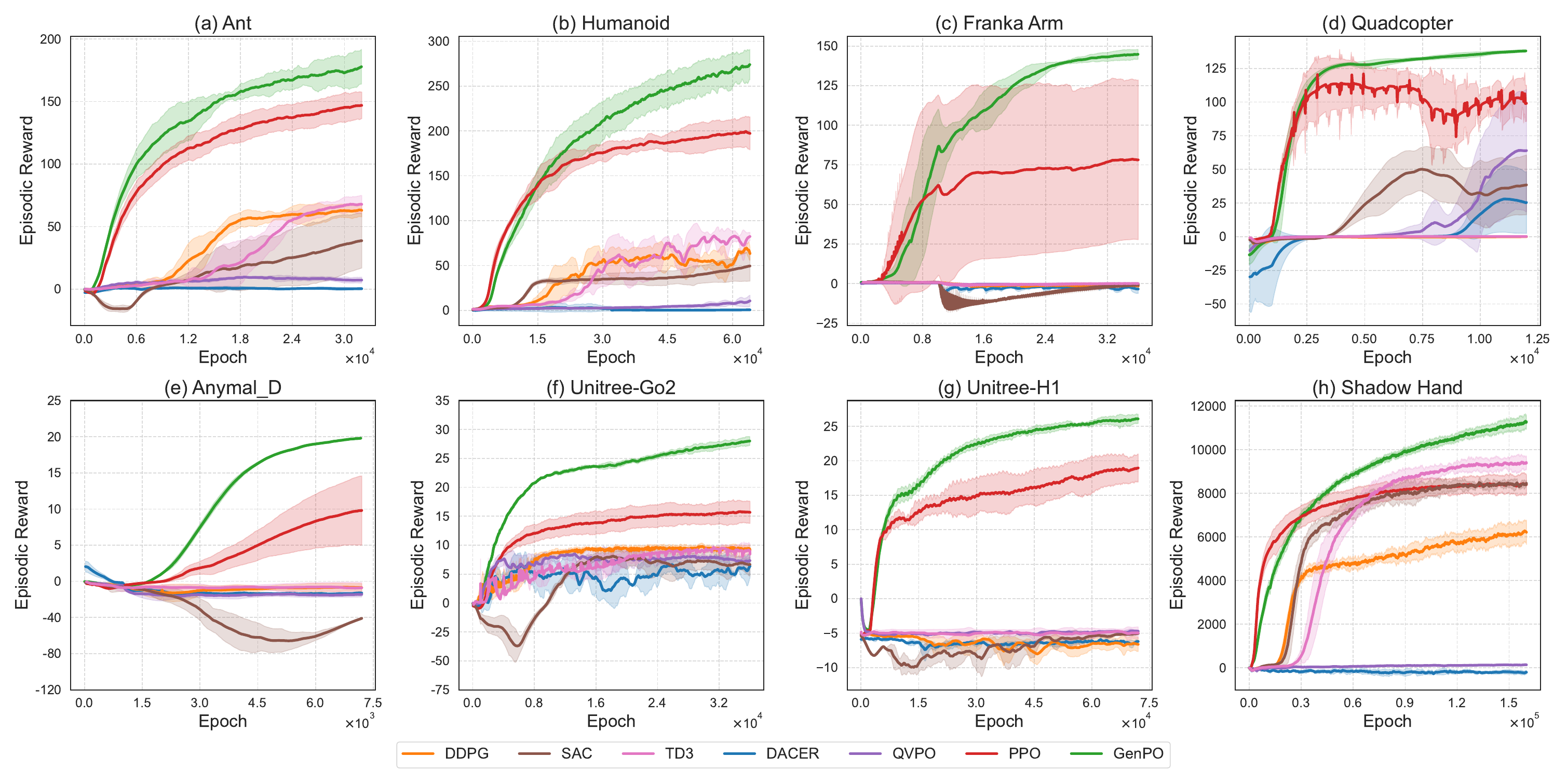}
    \caption{Learning curves across 8 IsaacLab benchmarks. Results are averaged over 5 runs. The x-axis denotes training epochs, and the y-axis shows average episodic return with one standard deviation shaded.}
    \label{fig:comparison}
\end{figure}
\begin{table}[ht]
 \vspace{-1mm}
 \caption{Comparison of the average final rewards of GenPO with some prevalent RL methods in IsaacLab Benchmarks. The maximum value for each task is shown in bold. The standard deviation of the five runs is enclosed in parentheses.}
 \resizebox{\linewidth}{!}{%
 \begin{tabular}{c||cccccccc}
 \toprule
 Algorithm & \makecell[c]{Ant} & \makecell[c]{Humanoid} & \makecell[c]{Franka-Arm} & \makecell[c]{Quadcopter} & \makecell[c]{Anymal-D} & \makecell[c]{Unitree-Go2} & \makecell[c]{Unitree-H1} & \makecell[c]{Shadow-Hand} \\
 \midrule
 DDPG~\cite{lillicrap2015continuous} & 62.96(5.18) & 63.34(7.39) & -1.61(0.76) & 0.03(0.13) & -0.86(0.55) & 9.09(0.81) & -6.63(1.04) & 6209.59(559.44)\\
 TD3~\cite{fujimoto2018addressing}  & 67.80(7.05) & 82.36(4.70) & 0.03(0.12) & 0.17(0.15) & -0.89(0.77) & 8.44(1.05) & -4.97(0.87) & 9386.64(314.37) \\
 SAC~\cite{haarnoja2018soft2}  & 38.68(21.84) & 49.52(16.89) & -1.10(0.95) & 38.46(22.20) & -5.12(0.14) & 6.67(2.02) & -5.05(0.18) & 8459.74(135.79) \\
 DACER~\cite{wang2024diffusion}  & 0.29(0.98) & 0.58(0.01) & -3.43(2.82) & 25.39(22.99) & -1.62(0.09) & 6.37(2.07) & -6.21(0.17) & -207.43(88.82) \\
 QVPO~\cite{ding2024diffusion}  &  7.19(2.32)& 10.59(6.30) & -0.12(0.14) & 63.99(45.42) & -1.81(0.36) &  7.34(0.62)& -4.72(0.16) &134.36(39.05)  \\
 PPO~\cite{schulman2017proximal}  & 146.94(10.61) & 197.25(18.26) & 78.14(50.43) & 99.08(13.49) & 9.80(4.78) & 15.67(1.92) & 18.97(2.02) & 8402.21(435.64) \\
 GenPO(*) & \textbf{177.90(13.87)} & \textbf{273.94(16.96)} & \textbf{144.78(3.10)} & \textbf{137.95(0.84)} & \textbf{19.80(0.16)} & \textbf{28.01(0.76)} & \textbf{26.09(0.68)} & \textbf{11282.35(322.94)} \\
 \bottomrule
 \end{tabular}%
 }
 \label{tab:result}
 \vspace{-1mm}
\end{table}

Figure~\ref{fig:comparison} presents the mean episodic return (solid line) and one standard deviation (shaded area). GenPO and PPO achieve near-optimal performance with fewer environment interactions compared to off-policy baselines, benefiting from synchronized data collection that enhances training stability and convergence speed. 
In contrast, off-policy methods struggle in large-scale parallel settings. This issue is particularly pronounced in diffusion-based algorithms, whose policies have difficulty tracking the rapidly shifting data distribution in the replay buffer. 
Additionally, GenPO's generative diffusion policy enhances exploration over standard Gaussian policies, contributing to more efficient and robust policy optimization. These factors together explain GenPO's superior performance across diverse tasks.

Table~\ref{tab:result} reports the average episodic return (with standard deviation) across eight benchmark tasks. GenPO consistently achieves the highest mean return across all environments, significantly outperforming all algorithms.
Notably, on the most challenging tasks, such as Unitree robots locomotion and Shadow Hand manipulation, GenPO demonstrates not only higher returns but also lower variance, indicating greater training stability.

\subsection{Ablation Study}
\begin{figure}
    \centering
    \includegraphics[width=1.0\linewidth]{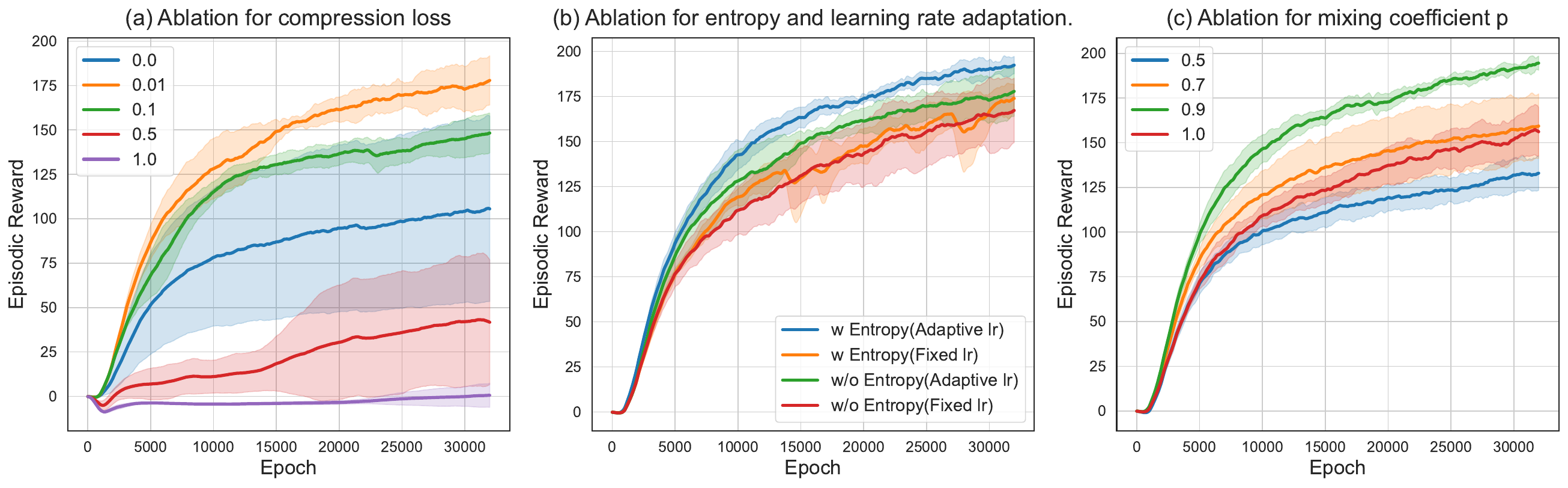}
    \caption{Ablation study results. (a) Effect of varying the compression loss coefficient $\nu$ on training stability and final performance. (b) Impact of entropy and learning rate adaptation on exploration and convergence. (c) Performance under different mixing coefficients $p$ in flow policies.}
    \label{fig:ablation}
\end{figure}
To better understand the significance of each component in GenPO and explain its superior performance on the IsaacLab benchmark, we conducted the ablation study focusing on three key design choices: 1) the effect of the compression loss in the training objective; 2) the impact of entropy and KL divergence estimation; and 3) the choice of the mixing coefficient $p$. We present results on the Isaac-Ant-v0 task as a representative example, as similar trends are observed across other environments. Additional ablation results are provided in the supplementary materials.

\textbf{Effect of Compression Loss.} To validate the impact of the compression loss on final performance, we vary its coefficient $\nu$ in (\ref{eq:loss}) and analyze the resulting training curves in Figure~\ref{fig:ablation}(a). We evaluate $\nu \in \{0, 0.01, 0.1, 0.5, 1.0\}$ and find that $\nu = 0.01$ provides the best balance between stability and regularization. Larger values overly constrain the policy and lead to poor performance, while excluding the loss ($\nu=0$) slows convergence. Based on this tradeoff, we set $\nu=0.01$ in all experiments.

\textbf{Entropy and Learning rate adaptation.} Figure~\ref{fig:ablation}(b) illustrates the effect of entropy regularization and learning rate adaptation.
Adaptive adjustment of the learning rate leads to faster and more stable convergence, while entropy regularization enhances exploration and yields higher returns.
Empirical results confirm that both components, implemented as described in Section~\ref{sect:practical}, substantially improve overall performance.
In all tasks, we employ an adaptive learning rate scheme and include an entropy regularization term in the loss defined in Eq.~(\ref{eq:loss}).

\textbf{Mixing coefficient.} As shown in Figure~\ref{fig:ablation}(c), GenPO performs best when the mixing coefficient is set to $p = 0.9$.
This is because setting $p$ too low will cause the forward mixing process (\ref{eq:fpo_forward}) to be numerically unstable, making the value too large and affecting the stability of training.
On the contrary, too high $p$ will introduce redundant exploration in the doubled dummy action space, thus slowing down the convergence.
Therefore, choosing $p = 0.9$ can achieve a practical trade-off between stability and exploration efficiency, so we set $p$ to 0.9 for all experiments.
\vspace{-3mm}
\section{Conclusion, Limitation and Future Works}
\label{sect:lim}
\vspace{-2mm}
This paper proposes Generative Diffusion Policy Optimization (GenPO), a novel approach that integrates generative diffusion policies into the on-policy RL framework. GenPO enables tractable log-likelihood computation of actions in the diffusion model, which is unavailable in prior works, and further unbiasedly estimates the entropy and KL divergence of the diffusion policy, thereby achieving entropy regularization and adaptive learning rate adjustment during policy update. Finally, GenPO achieves superior performance compared with existing RL baselines, including diffusion-based RL methods, on eight IsaacLab benchmarks, with comparable sample efficiency and faster convergence, highlighting the potential of generative diffusion policies in on-policy RL for large-scale simulation and real-world robotics control.

Despite its excellent performance, GenPO faces the problem of relatively high computational and memory overhead to be resolved in the future. While the GPU's parallelism helps to mitigate this issue, GenPO still incurs considerable complexity in computing the Jacobian determinant. Hence, our future work will focus on exploring how to optimize the computational and memory efficiency of GenPO to facilitate its deployment in more real-world applications.

\section*{Acknowledgement}
This work was supported by National Natural Science Foundation of China (62303319, 62406195), Shanghai Local College Capacity Building Program (23010503100), ShanghaiTech AI4S Initiative SHTAI4S202404, HPC Platform of ShanghaiTech University, and MoE Key Laboratory of Intelligent Perception and Human-Machine Collaboration (ShanghaiTech University), Shanghai Engineering Research Center of Intelligent Vision and Imaging. This work was also supported in part by computational resources provided by Fcloud CO., LTD. 

\bibliographystyle{plain}
\bibliography{reference}

\clearpage
\appendix

\section{Environmental Details}
% \subsection{Experimental Environment Introduction}
\begin{figure}[ht]
    \centering
    \includegraphics[width=1.0\linewidth]{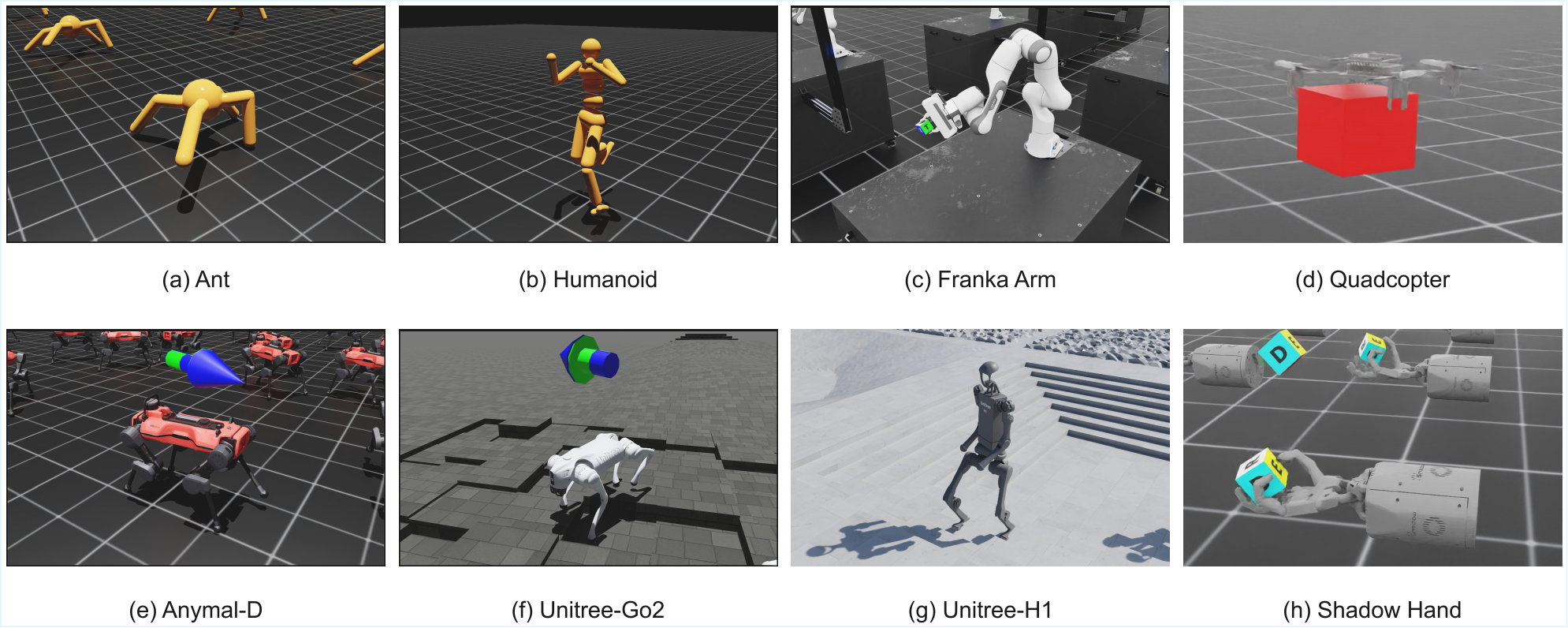}
    \caption{Eight Isaaclab benchmark visualizations, eight images from~\url{https://isaac-sim.github.io/IsaacLab/main/source/overview/environments.html}. From (a) to (h) are  Isaac-Ant-v0,  Isaac-Humanoid-v0, Isaac-Lift-Cube-Franka-v0, Isaac-Quadcopter-Direct-v0, Isaac-Velocity-Flat-Anymal-D-v0, Isaac-Velocity-Rough-Unitree-Go2-v0, Isaac-Velocity-Rough-H1-v0, and Isaac-Repose-Cube-Shadow-Direct-v0. 
    }
    \label{isaaclab}
\end{figure}
IsaacLab~\footnote{https://isaac-sim.github.io/IsaacLab/main/index.html} is a unified and modular framework for robot learning, designed to streamline common workflows in robotics research, including reinforcement learning, imitation learning, and motion planning. The framework leverages NVIDIA Isaac Sim for high-fidelity simulation and benefits from PhysX's GPU-accelerated physics engine as well as a tile-based rendering API for vectorized rendering.

As displayed in Figure~\ref{isaaclab}, we selected 8 representative and challenging environments, which can be roughly divided into the following categories according to the official manual of Isaaclab~\url{https://isaac-sim.github.io/IsaacLab/main/source/overview/environments.html}: 

\begin{enumerate}
    \item Classic: These classic tasks are derived from the IsaacGymEnv implementation of MuJoCo-style environments, providing standardized benchmarks for locomotion and control in continuous action spaces.
    \begin{itemize}
        \item \textbf{Isaac-Ant-v0.} The state and action spaces are $(\mathcal{S},\mathcal{A})\in \mathbb{R}^{60\times8}$. The task involves controlling a MuJoCo Ant robot to move in a specified direction.
        \item \textbf{Isaac-Humanoid-v0.} The state and action spaces are $(\mathcal{S},\mathcal{A})\in \mathbb{R}^{87\times21}$. The objective is to control a MuJoCo Humanoid robot to move in a specified direction.
    \end{itemize}
\item Manipulation: These environments involve manipulation tasks performed by a fixed-base robotic arm or a dexterous hand, such as object reaching, grasping, or in-hand rotation.
\begin{itemize}
    \item \textbf{Isaac-Lift-Cube-Franka-v0.} The state and action spaces are $(\mathcal{S},\mathcal{A})\in \mathbb{R}^{36\times8}$. The task involves controlling a Franka Emika robot arm to pick up a cube and transport it to a randomly sampled target position.
    \item \textbf{Isaac-Repose-Cube-Shadow-Direct-v0.} The state and action spaces are $(\mathcal{S},\mathcal{A})\in \mathbb{R}^{157\times20}$. The task requires using a Shadow Dexterous Hand to perform in-hand reorientation of a cube to match a target orientation.
\end{itemize}
\item Locomotion: This category includes legged locomotion environments that challenge agents to coordinate multiple degrees of freedom to achieve stable and directed movement. Tasks include forward walking, turning, and terrain adaptation, typically implemented with humanoid robots.
\begin{itemize}
    \item \textbf{Isaac-Velocity-Flat-Anymal-D-v0.} The state and action spaces are $(\mathcal{S},\mathcal{A})\in \mathbb{R}^{48\times12}$.  The task requires controlling an ANYmal D quadruped robot to follow a commanded velocity trajectory on flat terrain
    \item \textbf{Isaac-Velocity-Rough-Unitree-Go2-v0.} The state and action spaces are 
 $(\mathcal{S},\mathcal{A})\in \mathbb{R}^{235\times12}$. The task involves controlling a Unitree Go2 quadruped robot to follow a commanded velocity trajectory over rough terrain.
    \item \textbf{Isaac-Velocity-Rough-H1-v0.} The state and action spaces are $(\mathcal{S},\mathcal{A})\in \mathbb{R}^{256\times19}$. The task involves controlling a Unitree H1 humanoid robot to follow a commanded velocity trajectory over rough terrain.
\end{itemize}

\item Others:
\begin{itemize}
    \item \textbf{Isaac-Quadcopter-Direct-v0.} The state and action spaces are $(\mathcal{S},\mathcal{A})\in \mathbb{R}^{12\times4}$. The task is to control quadrotors to fly and hover at a designated goal position by applying thrust commands.
\end{itemize}
\end{enumerate}
Moreover, based on the IsaacLab Engine, the an anonymous link for the visualization of our experimental results is shown in~\url{anonymous-project365.github.io}.

\section{Training setups}
\subsection{Hardware Configurations}
All experiments were carried out on a server equipped with two Intel Xeon Gold 6430 CPUs (32 cores per socket, 64 threads total per CPU, 128 threads total), with a base frequency of 2.1 GHz and a maximum turbo frequency of 3.4 GHz. The system supports 52-bit physical and 57-bit virtual addressing. For GPU acceleration, we used 8 NVIDIA GeForce RTX 4090 D GPUs, each with 24 GB of GDDR6X memory, connected via PCIe. The GPUs support CUDA 12.8 and were operating under the NVIDIA driver version 570.124.04. The machine is configured with NUMA topology across 2 nodes, each handling 64 logical CPUs. No GPU MIG (Multi-Instance GPU) or ECC was enabled during the experiments.

\subsection{Reinforcement Learning Framework in IsaacLab}
IsaacLab provides native integration with several reinforcement learning libraries, including RSL-RL~\url{https://github.com/leggedrobotics/rsl_rl}, RL-Games~\url{https://github.com/Denys88/rl_games}, SKRL~\url{https://skrl.readthedocs.io/en/latest/}, and Stable-Baselines3~\url{https://stable-baselines3.readthedocs.io/en/master/index.html}, each of which exposes a distinct API for agent-environment interaction. In this work, we adopt SKRL as our primary framework for implementing baseline algorithms. SKRL is an open-source, modular, and extensible library that provides standardized implementations of widely used reinforcement learning methods. Specifically, we utilize its implementations of PPO~\url{https://skrl.readthedocs.io/en/latest/api/agents/ppo.html}, SAC~\url{https://skrl.readthedocs.io/en/latest/api/agents/sac.html}, TD3~\url{https://skrl.readthedocs.io/en/latest/api/agents/td3.html}, and DDPG~\url{https://skrl.readthedocs.io/en/latest/api/agents/ddpg.html} for baseline evaluation.
For algorithms such as QVPO and DACER, we design custom environment wrappers to adapt the existing interface without modifying the underlying simulation environments. Our proposed method, GenPO, is implemented using RSL-RL, a lightweight and high-performance framework tailored for robotics and continuous control, which emphasizes computational efficiency and ease of deployment.
\subsection{Hyperparameters}
Tables~\ref{tab:hyper_ant} and Table~\ref{tab:hyper_gen} summarize the hyperparameter configurations used across our experiments. For the baseline algorithms: PPO, SAC, TD3, and DDPG, we adopt the default hyperparameter settings provided by the SKRL library. For our proposed method, GenPO, we align its hyperparameter configuration with that of PPO to ensure a fair and controlled comparison.

\begin{table}[H]
\centering
\caption{Hyper-parameters used in the Isaaclab-Ant-v0.}
\resizebox{\textwidth}{!}{%
\begin{tabular}{llllllll}
\toprule
Hyperparameters               & GenPO & PPO & QVPO & DACER  & SAC  & TD3   & DDPG   \\
\midrule
hidden layers in actor network  &[400,200,100]  &[400,200,100]  & [256,256,256]     & [256,256,256]     & [400,200,100]     & [400,200,100]    & [400,200,100]     \\
hidden layers in critic network  &[400,200,100] &[400,200,100] & [256,256,256]   & [256,256,256]   & [400,200,100]   & [400,200,100]   & [400,200,100]   \\
Activation                     &   mish   & elu  & mish  & relu  & mish & mish & mish\\
Batch size                     &   4096   & 4096   & 4096   & 4096   & 4096  & 4096 & 4096   \\
Use GAE            & True   & True   & N/A   & N/A   & N/A  & N/A & N/A  \\
Discount for reward \(\gamma\) &   0.99   & 0.99  & 0.99  & 0.99  & 0.99 & 0.99 & 0.99 \\
GAE Smoothing Parameter \(\lambda\) & 0.95 & 0.95 &N/A &N/A &N/A &N/A &N/A \\
Learning rate for actor        &\(1\times 10^{-3}\)      & \(1 \times 10^{-3}\) & \(3 \times 10^{-4}\) & \(5 \times 10^{-4}\) & \(5 \times 10^{-4}\) & \(5 \times 10^{-4}\) & \(5 \times 10^{-4}\) \\
Learning rate for critic       &   \(1 \times 10^{-3}\)   & \(1 \times 10^{-3}\) & \(3 \times 10^{-4}\) & \(3 \times 10^{-4}\) & \(5 \times 10^{-4}\) & \(5 \times 10^{-4}\) & \(5 \times 10^{-4}\) \\
Actor Critic grad norm         &  1.0    & 1.0     & 1.0   & 1.0  &1.0   &1.0   & 1.0  \\
Memory size                    & N/A     & N/A & \(1 \times 10^{6}\) & \(1 \times 10^{6}\) & \(1 \times 10^{6}\) & \(1 \times 10^{6}\) & \(1 \times 10^{6}\) \\
Entropy coefficient            &  0.01  & 0.01   & N/A   & N/A   & N/A  & N/A & N/A  \\
Value loss coefficient         &  1.0  & 1.0   &  1.0  & 1.0   &1.0  & 1.0 & 1.0\\
Noise clip                     &  N/A & N/A   & N/A   & 0.5   & N/A  & N/A &N/A  \\
Surrogate clip &0.2&0.2&N/A&N/A&N/A&N/A&N/A \\
Diffusion steps            &  5   & N/A   & 20   &  20  & N/A  &  N/A &N/A \\
Desired KL & 0.01&0.01 &N/A &N/A&N/A&N/A&N/A \\
\bottomrule
\end{tabular}
}
\label{tab:hyper_ant}
\end{table}
\begin{table}[H]
\centering
\caption{Hyper-parameters used in the Isaaclab-Humanoid-v0.}
\resizebox{\textwidth}{!}{%
\begin{tabular}{llllllll}
\toprule
Hyperparameters               & GenPO & PPO & QVPO & DACER  & SAC  & TD3   & DDPG   \\
\midrule
hidden layers in actor network  &[400,200,100]  &[400,200,100]  & [256,256,256]     & [256,256,256]     & [400,200,100]     & [400,200,100]    & [400,200,100]     \\
hidden layers in critic network  &[400,200,100] &[400,200,100] & [256,256,256]   & [256,256,256]   & [400,200,100]   & [400,200,100]   & [400,200,100]   \\
Activation                     &   mish   & elu  & mish  & relu  & mish & mish & mish\\
Batch size                     &   4096   & 4096   & 4096   & 4096   & 4096  & 4096 & 4096   \\
Use GAE            & True   & True   & N/A   & N/A   & N/A  & N/A & N/A  \\
Discount for reward \(\gamma\) &   0.99   & 0.99  & 0.99  & 0.99  & 0.99 & 0.99 & 0.99 \\
GAE Smoothing Parameter \(\lambda\) & 0.95 & 0.95 &N/A &N/A &N/A &N/A &N/A \\
Learning rate for actor        &\(5\times 10^{-4}\)      & \(5 \times 10^{-4}\) & \(3 \times 10^{-4}\) & \(5 \times 10^{-4}\) & \(5 \times 10^{-4}\) & \(5 \times 10^{-4}\) & \(5 \times 10^{-4}\) \\
Learning rate for critic       &   \(5 \times 10^{-4}\)   & \(5 \times 10^{-4}\) & \(3 \times 10^{-4}\) & \(3 \times 10^{-4}\) & \(5 \times 10^{-4}\) & \(5 \times 10^{-4}\) & \(5 \times 10^{-4}\) \\
Actor Critic grad norm         &  1.0    & 1.0     & 1.0   & 1.0  &1.0   &1.0   & 1.0  \\
Memory size                    & N/A     & N/A & \(1 \times 10^{6}\) & \(1 \times 10^{6}\) & \(1 \times 10^{6}\) & \(1 \times 10^{6}\) & \(1 \times 10^{6}\) \\
Entropy coefficient            &  0.01  & 0.01   & N/A   & N/A   & N/A  & N/A & N/A  \\
Value loss coefficient         &  1.0  & 1.0   &  1.0  & 1.0   &1.0  & 1.0 & 1.0\\
Noise clip                     &  N/A & N/A   & N/A   & 0.5   & N/A  & N/A &N/A  \\
Surrogate clip &0.2&0.2&N/A&N/A&N/A&N/A&N/A \\
Diffusion steps            &  5   & N/A   & 20   &  20  & N/A  &  N/A &N/A \\
Desired KL & 0.01&0.01 &N/A &N/A&N/A&N/A&N/A \\
\bottomrule
\end{tabular}
}
\label{tab:hyper_humanoid}
\end{table}
\begin{table}[H]
\centering
\caption{Hyper-parameters used in the Isaac-Lift-Cube-Franka-v0.}
\resizebox{\textwidth}{!}{%
\begin{tabular}{llllllll}
\toprule
Hyperparameters               & GenPO & PPO & QVPO & DACER  & SAC  & TD3   & DDPG   \\
\midrule
hidden layers in actor network  &[256,128,64]  &[256,128,64]  & [256,256,256]     & [256,256,256]     & [256, 128, 64]     & [256, 128, 64]    & [256, 128, 64]     \\
hidden layers in critic network  &[256,128,64] &[256,128,64] & [256,256,256]   & [256,256,256]   & [256, 128, 64]   & [256, 128, 64]   & [256, 128, 64]  \\
Activation                     &   mish   & elu  & mish  & relu  & mish & mish & mish\\
Batch size                     &   4096   & 4096   & 4096   & 4096   & 4096  & 4096 & 4096   \\
Use GAE            & True   & True   & N/A   & N/A   & N/A  & N/A & N/A  \\
Discount for reward \(\gamma\) &   0.98   & 0.98  & 0.99  & 0.99  & 0.99 & 0.99 & 0.99 \\
GAE Smoothing Parameter \(\lambda\) & 0.95 & 0.95 &N/A &N/A &N/A &N/A &N/A \\
Learning rate for actor        &\(1\times 10^{-4}\)      & \(1 \times 10^{-4}\) & \(3 \times 10^{-4}\) & \(5 \times 10^{-4}\) & \(1 \times 10^{-4}\) & \(1 \times 10^{-4}\) & \(1 \times 10^{-4}\) \\
Learning rate for critic       &   \(1 \times 10^{-4}\)   & \(1 \times 10^{-4}\) & \(3 \times 10^{-4}\) & \(3 \times 10^{-4}\) & \(1 \times 10^{-4}\) & \(1 \times 10^{-4}\) & \(1 \times 10^{-4}\) \\
Actor Critic grad norm         &  1.0    & 1.0     & 1.0   & 1.0  &1.0   &1.0   & 1.0  \\
Memory size                    & N/A     & N/A & \(1 \times 10^{6}\) & \(1 \times 10^{6}\) & \(1 \times 10^{6}\) & \(1 \times 10^{6}\) & \(1 \times 10^{6}\) \\
Entropy coefficient            &  0.006  & 0.006   & N/A   & N/A   & N/A  & N/A & N/A  \\
Value loss coefficient         &  1.0  & 1.0   &  1.0  & 1.0   &1.0  & 1.0 & 1.0\\
Noise clip                     &  N/A & N/A   & N/A   & 0.5   & N/A  & N/A &N/A  \\
Surrogate clip &0.2&0.2&N/A&N/A&N/A&N/A&N/A \\
Diffusion steps            &  5   & N/A   & 20   &  20  & N/A  &  N/A &N/A \\
Desired KL & 0.01&0.01 &N/A &N/A&N/A&N/A&N/A \\
\bottomrule
\end{tabular}
}
\label{tab:hyper_franka}
\end{table}
\begin{table}[H]
\centering
\caption{Hyper-parameters used in the Isaac-Quadcopter-Direct-v0.}
\resizebox{\textwidth}{!}{%
\begin{tabular}{llllllll}
\toprule
Hyperparameters               & GenPO & PPO & QVPO & DACER  & SAC  & TD3   & DDPG   \\
\midrule
hidden layers in actor network  &[64,64]  &[64,64]  & [256,256,256]     & [256,256,256]     & [64, 64]     & [64, 64]    & [64, 64]     \\
hidden layers in critic network  &[64,64] &[64,64] & [256,256,256]   & [256,256,256]   & [64, 64]   & [64, 64]   & [64, 64]   \\
Activation                     &   mish   & elu  & mish  & relu  & mish & mish & mish\\
Batch size                     &   4096   & 4096   & 4096   & 4096   & 4096  & 4096 & 4096   \\
Use GAE            & True   & True   & N/A   & N/A   & N/A  & N/A & N/A  \\
Discount for reward \(\gamma\) &   0.99   & 0.99  & 0.99  & 0.99  & 0.99 & 0.99 & 0.99 \\
GAE Smoothing Parameter \(\lambda\) & 0.95 & 0.95 &N/A &N/A &N/A &N/A &N/A \\
Learning rate for actor        &\(5\times 10^{-4}\)      & \(5 \times 10^{-4}\) & \(3 \times 10^{-4}\) & \(5 \times 10^{-4}\) & \(5 \times 10^{-4}\) & \(5 \times 10^{-4}\) & \(5 \times 10^{-4}\) \\
Learning rate for critic       &   \(5 \times 10^{-4}\)   & \(5 \times 10^{-4}\) & \(3 \times 10^{-4}\) & \(3 \times 10^{-4}\) & \(5 \times 10^{-4}\) & \(5 \times 10^{-4}\) & \(5 \times 10^{-4}\) \\
Actor Critic grad norm         &  1.0    & 1.0     & 1.0   & 1.0  &1.0   &1.0   & 1.0  \\
Memory size                    & N/A     & N/A & \(1 \times 10^{6}\) & \(1 \times 10^{6}\) & \(1 \times 10^{6}\) & \(1 \times 10^{6}\) & \(1 \times 10^{6}\) \\
Entropy coefficient            &  0.01  & 0.01   & N/A   & N/A   & N/A  & N/A & N/A  \\
Value loss coefficient         &  1.0  & 1.0   &  1.0  & 1.0   &1.0  & 1.0 & 1.0\\
Noise clip                     &  N/A & N/A   & N/A   & 0.5   & N/A  & N/A &N/A  \\
Surrogate clip &0.2&0.2&N/A&N/A&N/A&N/A&N/A \\
Diffusion steps            &  5   & N/A   & 20   &  20  & N/A  &  N/A &N/A \\
Desired KL & 0.01&0.01 &N/A &N/A&N/A&N/A&N/A \\
\bottomrule
\end{tabular}
}
\label{tab:hyper_quadcopter}
\end{table}
\begin{table}[H]
\centering
\caption{Hyper-parameters used in the Isaac-Velocity-Flat-Anymal-D-v0.}
\resizebox{\textwidth}{!}{%
\begin{tabular}{llllllll}
\toprule
Hyperparameters               & GenPO & PPO & QVPO & DACER  & SAC  & TD3   & DDPG   \\
\midrule
hidden layers in actor network  &[128,128,128]  &[128,128,128]  & [256,256,256]     & [256,256,256]     & [128, 128, 128]   & [128, 128, 128]  & [128, 128, 128]    \\
hidden layers in critic network  &[128,128,128] &[128,128,128] & [256,256,256]   & [256,256,256]   & [128, 128, 128]   & [128, 128, 128]   & [128, 128, 128]  \\
Activation                     &   mish   & elu  & mish  & relu  & mish & mish & mish\\
Batch size                     &   4096   & 4096   & 4096   & 4096   & 4096  & 4096 & 4096   \\
Use GAE            & True   & True   & N/A   & N/A   & N/A  & N/A & N/A  \\
Discount for reward \(\gamma\) &   0.99   & 0.99  & 0.99  & 0.99  & 0.99 & 0.99 & 0.99 \\
GAE Smoothing Parameter \(\lambda\) & 0.95 & 0.95 &N/A &N/A &N/A &N/A &N/A \\
Learning rate for actor        &\(1\times 10^{-3}\)      & \(1 \times 10^{-3}\) & \(3 \times 10^{-4}\) & \(5 \times 10^{-4}\) & \(1 \times 10^{-3}\) & \(1 \times 10^{-3}\) & \(1 \times 10^{-3}\) \\
Learning rate for critic       &   \(1 \times 10^{-3}\)   & \(1 \times 10^{-3}\) & \(3 \times 10^{-4}\) & \(3 \times 10^{-4}\) & \(1 \times 10^{-3}\) & \(1 \times 10^{-3}\) & \(1 \times 10^{-3}\) \\
Actor Critic grad norm         &  1.0    & 1.0     & 1.0   & 1.0  &1.0   &1.0   & 1.0  \\
Memory size                    & N/A     & N/A & \(1 \times 10^{6}\) & \(1 \times 10^{6}\) & \(1 \times 10^{6}\) & \(1 \times 10^{6}\) & \(1 \times 10^{6}\) \\
Entropy coefficient            &  0.005  & 0.005   & N/A   & N/A   & N/A  & N/A & N/A  \\
Value loss coefficient         &  1.0  & 1.0   &  1.0  & 1.0   &1.0  & 1.0 & 1.0\\
Noise clip                     &  N/A & N/A   & N/A   & 0.5   & N/A  & N/A &N/A  \\
Surrogate clip &0.2&0.2&N/A&N/A&N/A&N/A&N/A \\
Diffusion steps            &  5   & N/A   & 20   &  20  & N/A  &  N/A &N/A \\
Desired KL & 0.01&0.01 &N/A &N/A&N/A&N/A&N/A\\
\bottomrule
\end{tabular}
}
\label{tab:hyper_anymal-d}
\end{table}
\begin{table}[H]
\centering
\caption{Hyper-parameters used in the Isaac-Velocity-Rough-Unitree-Go2-v0.}
\resizebox{\textwidth}{!}{%
\begin{tabular}{llllllll}
\toprule
Hyperparameters               & GenPO & PPO & QVPO & DACER  & SAC  & TD3   & DDPG   \\
\midrule
hidden layers in actor network  &[512, 256, 128]  &[512, 256, 128]  & [256,256,256]     & [256,256,256]     & [512, 256, 128]     & [512, 256, 128]    & [512, 256, 128]     \\
hidden layers in critic network  &[512, 256, 128] &[512, 256, 128] & [256,256,256]   & [256,256,256]   & [512, 256, 128]   & [512, 256, 128]   & [512, 256, 128]   \\
Activation                     &   mish   & elu  & mish  & relu  & mish & mish & mish\\
Batch size                     &   4096   & 4096   & 4096   & 4096   & 4096  & 4096 & 4096   \\
Use GAE            & True   & True   & N/A   & N/A   & N/A  & N/A & N/A  \\
Discount for reward \(\gamma\) &   0.99   & 0.99  & 0.99  & 0.99  & 0.99 & 0.99 & 0.99 \\
GAE Smoothing Parameter \(\lambda\) & 0.95 & 0.95 &N/A &N/A &N/A &N/A &N/A \\
Learning rate for actor        &\(1\times 10^{-3}\)      & \(1 \times 10^{-3}\) & \(3 \times 10^{-4}\) & \(5 \times 10^{-4}\) & \(1 \times 10^{-3}\) & \(1 \times 10^{-3}\) & \(1 \times 10^{-3}\) \\
Learning rate for critic       &   \(1 \times 10^{-3}\)   & \(1 \times 10^{-3}\) & \(3 \times 10^{-4}\) & \(3 \times 10^{-4}\) & \(1 \times 10^{-3}\) & \(1 \times 10^{-3}\) & \(1 \times 10^{-3}\) \\
Actor Critic grad norm         &  1.0    & 1.0     & 1.0   & 1.0  &1.0   &1.0   & 1.0  \\
Memory size                    & N/A     & N/A & \(1 \times 10^{6}\) & \(1 \times 10^{6}\) & \(1 \times 10^{6}\) & \(1 \times 10^{6}\) & \(1 \times 10^{6}\) \\
Entropy coefficient            &  0.01  & 0.01   & N/A   & N/A   & N/A  & N/A & N/A  \\
Value loss coefficient         &  1.0  & 1.0   &  1.0  & 1.0   &1.0  & 1.0 & 1.0\\
Noise clip                     &  N/A & N/A   & N/A   & 0.5   & N/A  & N/A &N/A  \\
Surrogate clip &0.2&0.2&N/A&N/A&N/A&N/A&N/A \\
Diffusion steps            &  5   & N/A   & 20   &  20  & N/A  &  N/A &N/A \\
Desired KL & 0.01&0.01 &N/A &N/A&N/A&N/A&N/A \\
\bottomrule
\end{tabular}
}
\label{tab:hyper_go2}
\end{table}
\begin{table}[H]
\centering
\caption{Hyper-parameters used in the Isaac-Velocity-Rough-H1-v0.}
\resizebox{\textwidth}{!}{%
\begin{tabular}{llllllll}
\toprule
Hyperparameters               & GenPO & PPO & QVPO & DACER  & SAC  & TD3   & DDPG   \\
\midrule
hidden layers in actor network  &[512, 256, 128]  &[512, 256, 128]  & [256,256,256]     & [256,256,256]     & [512, 256, 128]     & [512, 256, 128]    & [512, 256, 128]     \\
hidden layers in critic network  &[512, 256, 128] &[512, 256, 128] & [256,256,256]   & [256,256,256]   & [512, 256, 128]   & [512, 256, 128]   & [512, 256, 128]   \\
Activation                     &   mish   & elu  & mish  & relu  & mish & mish & mish\\
Batch size                     &   4096   & 4096   & 4096   & 4096   & 4096  & 4096 & 4096   \\
Use GAE            & True   & True   & N/A   & N/A   & N/A  & N/A & N/A  \\
Discount for reward \(\gamma\) &   0.99   & 0.99  & 0.99  & 0.99  & 0.99 & 0.99 & 0.99 \\
GAE Smoothing Parameter \(\lambda\) & 0.95 & 0.95 &N/A &N/A &N/A &N/A &N/A \\
Learning rate for actor        &\(1\times 10^{-3}\)      & \(1 \times 10^{-3}\) & \(3 \times 10^{-4}\) & \(5 \times 10^{-4}\) & \(1 \times 10^{-3}\) & \(1 \times 10^{-3}\) & \(1 \times 10^{-3}\) \\
Learning rate for critic       &   \(1 \times 10^{-3}\)   & \(1 \times 10^{-3}\) & \(3 \times 10^{-4}\) & \(3 \times 10^{-4}\) & \(1 \times 10^{-3}\) & \(1 \times 10^{-3}\) & \(1 \times 10^{-3}\) \\
Actor Critic grad norm         &  1.0    & 1.0     & 1.0   & 1.0  &1.0   &1.0   & 1.0  \\
Memory size                    & N/A     & N/A & \(1 \times 10^{6}\) & \(1 \times 10^{6}\) & \(1 \times 10^{6}\) & \(1 \times 10^{6}\) & \(1 \times 10^{6}\) \\
Entropy coefficient            &  0.01  & 0.01   & N/A   & N/A   & N/A  & N/A & N/A  \\
Value loss coefficient         &  1.0  & 1.0   &  1.0  & 1.0   &1.0  & 1.0 & 1.0\\
Noise clip                     &  N/A & N/A   & N/A   & 0.5   & N/A  & N/A &N/A  \\
Surrogate clip &0.2&0.2&N/A&N/A&N/A&N/A&N/A \\
Diffusion steps            &  5   & N/A   & 20   &  20  & N/A  &  N/A &N/A \\
Desired KL & 0.01&0.01 &N/A &N/A&N/A&N/A&N/A \\
\bottomrule
\end{tabular}
}
\label{tab:hyper_h1}
\end{table}
\begin{table}[H]
\centering
\caption{Hyper-parameters used in the Isaac-Repose-Cube-Shadow-Direct-v0.}
\resizebox{\textwidth}{!}{%
\begin{tabular}{llllllll}
\toprule
Hyperparameters               & GenPO & PPO & QVPO & DACER  & SAC  & TD3   & DDPG   \\
\midrule
hidden layers in actor network  &[512, 512, 256, 128]  &[512, 512, 256, 128]  & [256,256,256]     & [256,256,256]     & [512, 512, 256, 128]     & [512, 512, 256, 128]    & [512, 512, 256, 128]     \\
hidden layers in critic network  &[512, 512, 256, 128] &[512, 512, 256, 128] & [256,256,256]   & [256,256,256]   & [512, 512, 256, 128]   & [512, 512, 256, 128]   & [512, 512, 256, 128]   \\
Activation                     &   mish   & elu  & mish  & relu  & mish & mish & mish\\
Batch size                     &   4096   & 4096   & 4096   & 4096   & 4096  & 4096 & 4096   \\
Use GAE            & True   & True   & N/A   & N/A   & N/A  & N/A & N/A  \\
Discount for reward \(\gamma\) &   0.99   & 0.99  & 0.99  & 0.99  & 0.99 & 0.99 & 0.99 \\
GAE Smoothing Parameter \(\lambda\) & 0.95 & 0.95 &N/A &N/A &N/A &N/A &N/A \\
Learning rate for actor        &\(5\times 10^{-4}\)      & \(5 \times 10^{-4}\) & \(3 \times 10^{-4}\) & \(5 \times 10^{-4}\) & \(5 \times 10^{-4}\) & \(5 \times 10^{-4}\) & \(5 \times 10^{-4}\) \\
Learning rate for critic       &   \(5 \times 10^{-4}\)   & \(5 \times 10^{-4}\) & \(3 \times 10^{-4}\) & \(3 \times 10^{-4}\) & \(5 \times 10^{-4}\) & \(5 \times 10^{-4}\) & \(5 \times 10^{-4}\) \\
Actor Critic grad norm         &  1.0    & 1.0     & 1.0   & 1.0  &1.0   &1.0   & 1.0  \\
Memory size                    & N/A     & N/A & \(1 \times 10^{6}\) & \(1 \times 10^{6}\) & \(1 \times 10^{6}\) & \(1 \times 10^{6}\) & \(1 \times 10^{6}\) \\
Entropy coefficient            &  0.005  & 0.005   & N/A   & N/A   & N/A  & N/A & N/A  \\
Value loss coefficient         &  1.0  & 1.0   &  1.0  & 1.0   &1.0  & 1.0 & 1.0\\
Noise clip                     &  N/A & N/A   & N/A   & 0.5   & N/A  & N/A &N/A  \\
Surrogate clip &0.2&0.2&N/A&N/A&N/A&N/A&N/A \\
Diffusion steps            &  5   & N/A   & 20   &  20  & N/A  &  N/A &N/A \\
Desired KL & 0.016&0.016 &N/A &N/A&N/A&N/A&N/A \\
\bottomrule
\end{tabular}
}
\label{tab:hyper_hand}
\end{table}
\begin{table}[H]
\centering
\caption{Hyper-parameters used in GenPO.}
\resizebox{\textwidth}{!}{%
\begin{tabular}{c||cccccccc}
\toprule
Hyperparameter & Ant & Humanoid & Franka Arm & Quadcopter & Anymal-D& Unitree-Go2&Unitree-H1&Shadow Hand \\
\midrule
  Compress coefficient $\lambda$  &  0.01      &  0.01 &  0.01  &  0.01  &  0.01 &  0.01 &  0.01    &  0.01        \\
  Time embedding dimension     & 32   &       32         &    32       &      16       &   32   &32&32&32       \\
  Hidden layers in time embedding & [256,256]  & [256,256]  & [256,256]  & [128,128]  & [256,256] & [256,256] & [256,256] & [256,256]      \\
  Mixing coefficient $p$     &  0.9  &  0.9  &  0.9 &  0.9 &  0.9 &  0.9 &  0.9 &  0.9    \\

\bottomrule
\end{tabular}%
}
\label{tab:hyper_gen}
\end{table}
\section{Details of Model Architecture}
\subsection{Sinusoidal Positional Embedding} During training, instead of directly concatenating the time step $t$, state $s_t$, and action $a_t$ as raw input features, we incorporate sinusoidal positional embeddings to encode the temporal component. This approach, inspired by the positional encoding technique commonly used in denoising diffusion probabilistic models (DDPMs), maps each timestep $ t \in \{0, 1, \ldots, T\}$to a fixed-dimensional vector $ e_t \in \mathbb{R}^d $ as follows:

\begin{equation}
    e_t^{(2i)} = \sin \left( \frac{t}{10000^{2i/d}} \right), \quad e_t^{(2i+1)} = \cos \left( \frac{t}{10000^{2i/d}} \right), \quad \text{for } i = 0, 1, \ldots, \frac{d}{2} - 1.
\end{equation}

The resulting embedding $e_t$ is then passed through a multi-layer perceptron (MLP) to allow for nonlinear transformation and projection into the model's feature space. The MLP output is subsequently concatenated with the state $s_t$ and action $a_t$ to form the final input representation:
\begin{equation}
    \tilde{x}_t = \text{concat}(s_t, a_t, \text{MLP}(e_t)),
\end{equation}
which enriches the model input with a smooth, continuous representation of temporal progression. Unlike raw scalar timestep concatenation, sinusoidal embeddings provide a structured encoding that helps the model infer temporal relationships and ordering, even in the absence of explicit recurrence or attention mechanisms.

\subsection{Actor and Critic Model Architecture} In the Isaac-Ant-v0 environment, the actor network, corresponding to the flow policy, is implemented as a multi-layer perceptron (MLP) with hidden layer dimensions [400,200,100]. The activation function used is Mish. The input to the actor consists of the concatenation of the current state, action, and sinusoidal time embedding vectors, resulting in an input dimension of 
$|\mathcal{A}|+|\mathcal{S}|+|\mathcal{T}|$, The output dimension is equal to the action dimension $|\mathcal{A}|$, representing the flow-based action refinement.
The critic network is also implemented as an MLP with the same hidden layer structure 
[400,200,100] and Mish activation. It receives only the current state as input and outputs a scalar value corresponding to the estimated state value.

\section{Additional Experiments}
\subsection{Effect of Different Parallel Environments Size} To assess the effect of parallelization on the learning efficiency of GenPO, we conducted an experiment varying the number of parallel environments. We trained GenPO agents using 512, 1024, 2048, 4096, and 8192 parallel environments, respectively. All other hyperparameters were kept consistent across these runs. 

\begin{figure}[ht]
    \centering
    \includegraphics[width=1.0\linewidth]{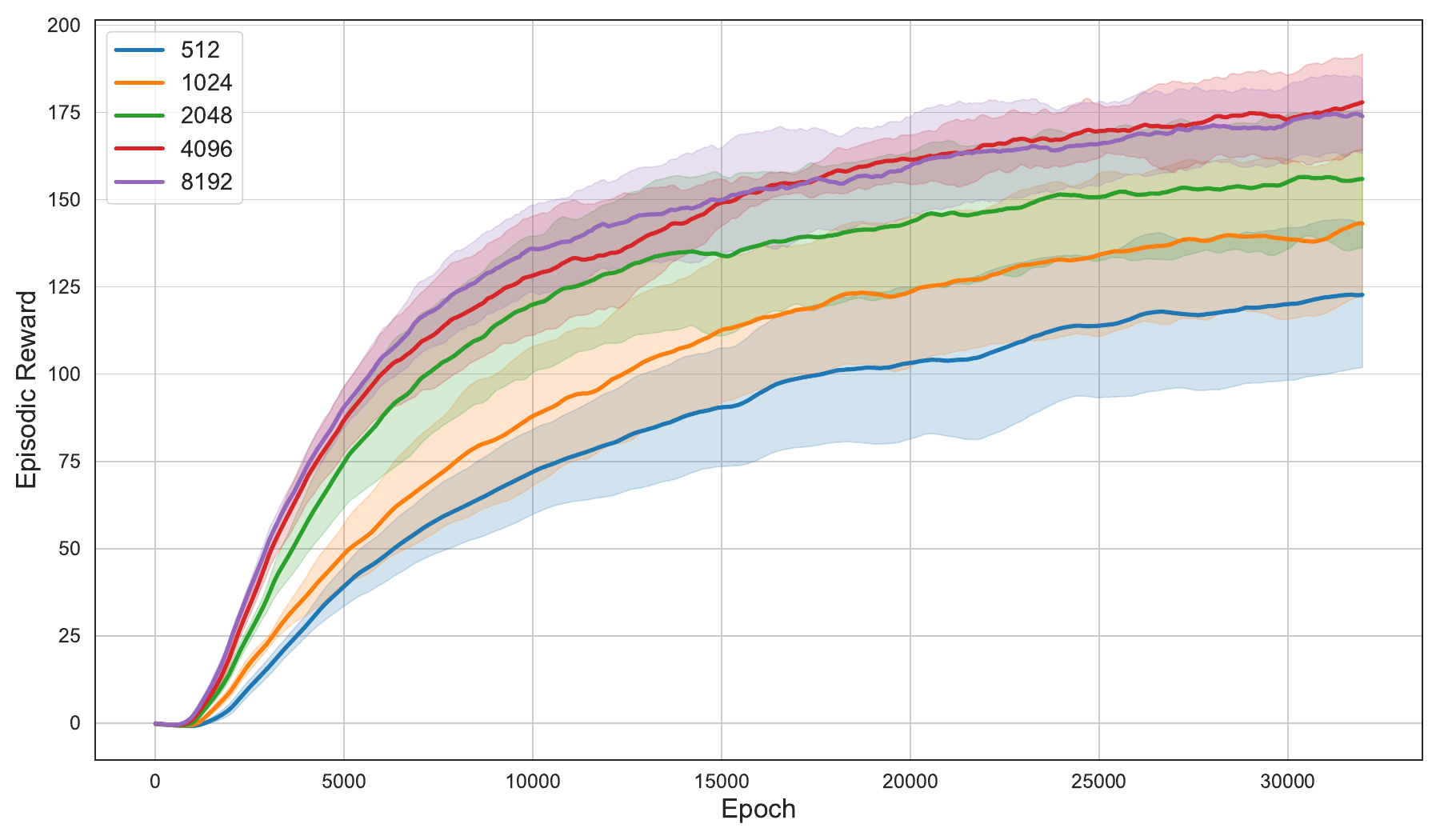}
    \caption{Learning curves for the Isaaclab-Ant-v0 benchmark with different numbers of parallel environments. Results are averaged over 5 runs. The x-axis denotes training epochs, and the y-axis shows average episodic return with one standard deviation shaded.}
    \label{fig:diff_num_envs}
\end{figure}

As illustrated in Figure~\ref{fig:diff_num_envs}, increasing the number of parallel environments generally leads to improved learning performance. Agents trained with a larger number of environments exhibit faster convergence and achieve higher final rewards compared to those trained with fewer environments. However, the performance gain from using 8192 environments over 4096 is marginal. To balance performance and computational efficiency, we fix the number of parallel environments to 4096 in all experiments.

\subsection{Effect of Different Flow Policy Steps} 
\begin{figure}[ht]
    \centering
    \includegraphics[width=1.0\linewidth]{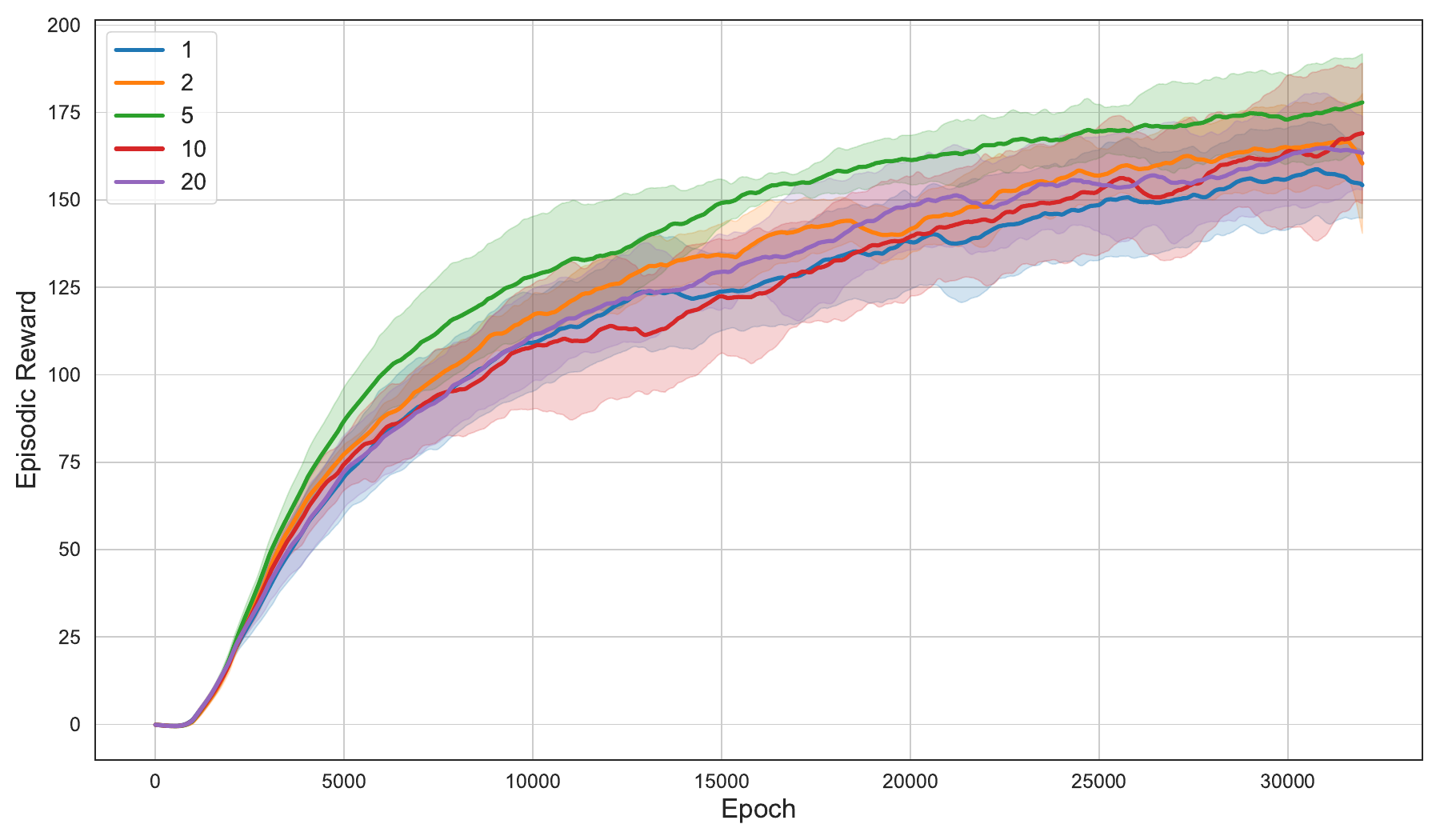}
    \caption{Learning curves for different flow policy time steps on the Isaaclab-Ant-v0 benchmark. Results are averaged over 5 runs.  The x-axis denotes training epochs, and the y-axis shows average episodic return with one standard deviation shaded.}
    \label{fig:diffusion_steps}
\end{figure}

We investigate the impact of the number of flow steps, denoted by 
$T$, on learning performance. To this end, we train agents using different values of $T\in\{1,2,5,10,20\}$, while keeping all other hyperparameters fixed. As shown in Figure~\ref{fig:diffusion_steps}, the number of flow steps influences convergence speed and performance stability. Notably, as long as $T$ is not too small (e.g., $T\geq2$), the overall performance remains robust.

\subsection{Effect of Different Operations on Dummy Action Space} 
\begin{figure}[ht]
    \centering
    \includegraphics[width=1.0\linewidth]{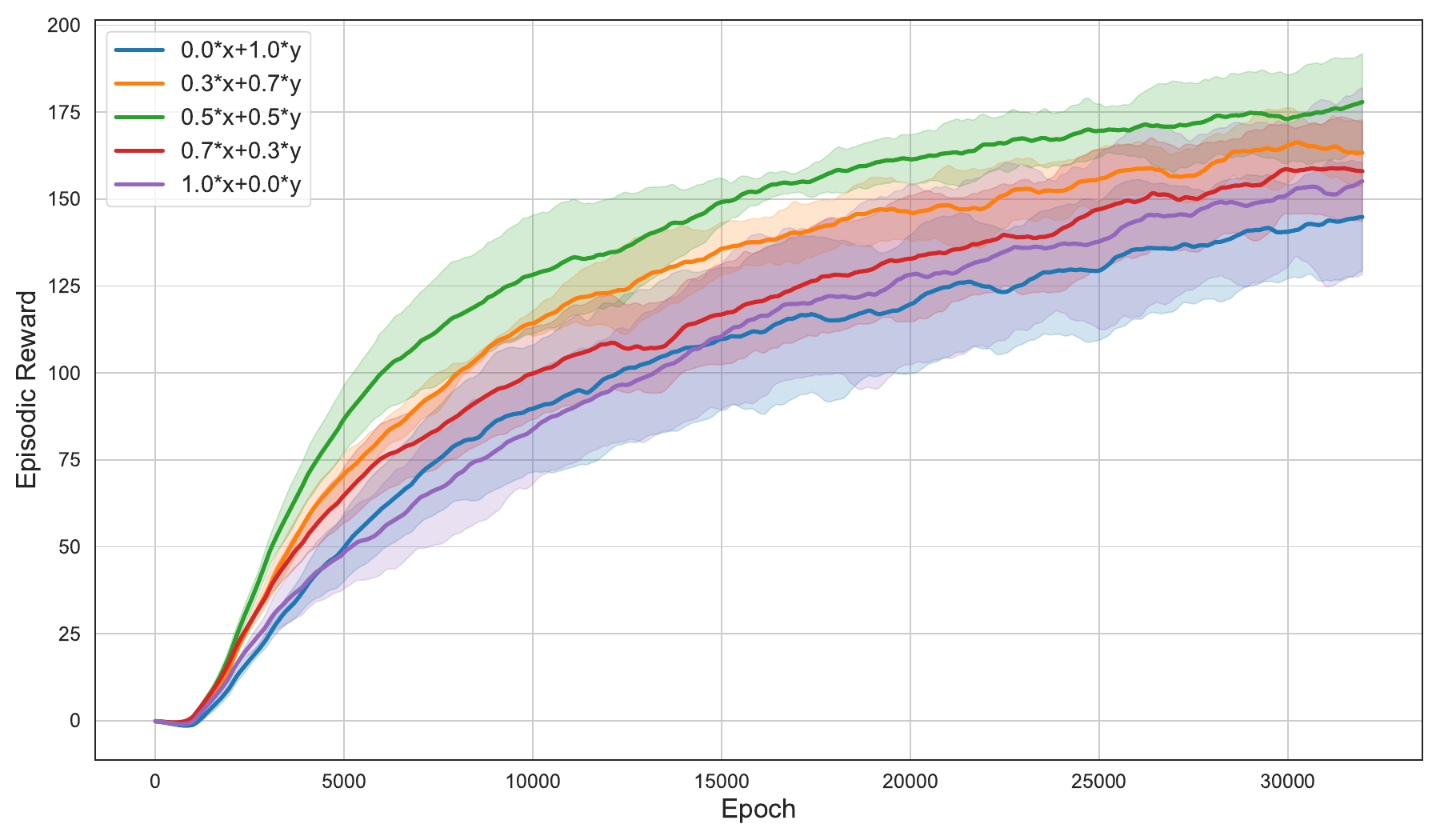}
    \caption{Learning curves for different operations on dummy action space on the Isaaclab-Ant-v0 benchmark. Results are averaged over 5 runs.  The x-axis denotes training epochs, and the y-axis shows average episodic return with one standard deviation shaded.}
    \label{fig:inter_env}
\end{figure}

We study the impact of recovering a single action from the dummy action $\tilde{a} = (x, y)$ by interpolating between the two partial components $x$, $y$. Specifically, we construct the real action as $\alpha x+(1-\alpha)\ y$ and $\alpha\in\{ 0.0,0.3,0.5,0.7,1.0\}$ respectively. As illustrated in Figure~\ref{fig:inter_env}, the best performance is achieved when $\alpha=0.5$. This observation highlights the benefit of equally leveraging both components of the dummy action. When $\alpha=0.0$ or $\alpha=1.0$, only one side of the dummy action space is utilized, leading to limited exploration and suboptimal performance. In contrast, interpolating between both components enhances exploration diversity and improves sample efficiency, which is crucial for effective policy optimization in high-dimensional or multimodal action spaces.
\subsection{Effect of Time Steps Embedding} 
\begin{figure}[ht]
    \centering
    \includegraphics[width=1.0\linewidth]{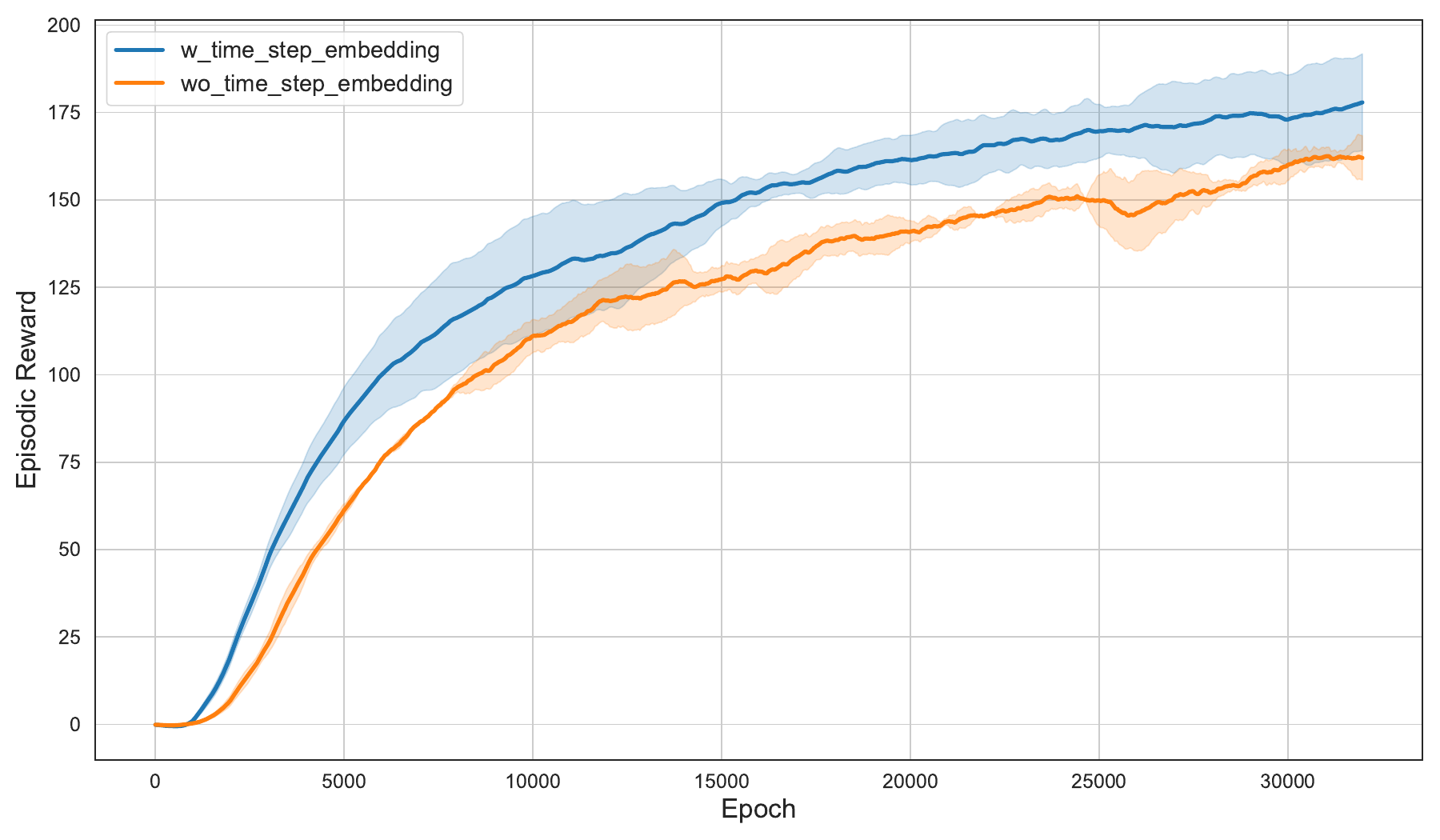}
    \caption{Learning curves on the Isaaclab-Ant-v0 benchmark with or without time step sinusoidal positional embedding. Results are averaged over 5 runs.  The x-axis denotes training epochs, and the y-axis shows average episodic return with one standard deviation shaded.}
    \label{fig:time_emb}
\end{figure}

We investigate the effect of sinusoidal time embeddings on the performance of the flow policy. As shown in Figure~\ref{fig:time_emb}, incorporating sinusoidal embeddings significantly improves the policy's ability to utilize temporal information across different stages of the diffusion process. The structured, high-dimensional representation provided by the sinusoidal encoding facilitates better discrimination of time steps, which in turn enhances the model's capacity to learn effective denoising functions.
In contrast, using simple concatenation of the scalar timestep with the state and action fails to provide sufficient temporal structure, resulting in degraded performance. Based on these observations, we adopt sinusoidal positional embeddings to encode time steps in all subsequent experiments.
\subsection{Training and Inference Time} 
\begin{figure}[ht]
    \centering
    \includegraphics[width=1.0\linewidth]{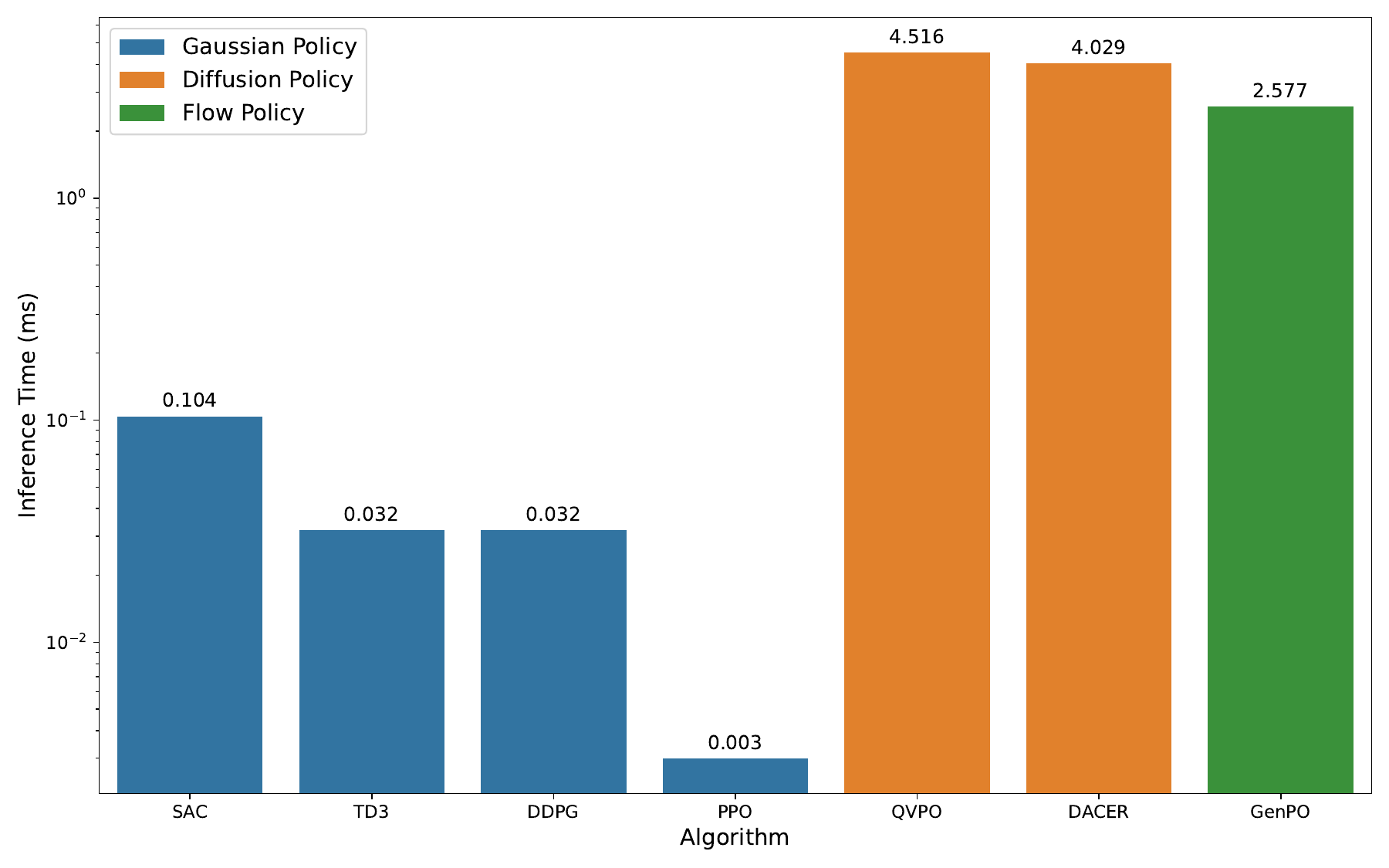}
    \caption{Inference time comparison on Issac-Ant-v0 environment.}
    \label{fig:inference}
\end{figure}

\begin{figure}[ht]
    \centering
    \includegraphics[width=1.0\linewidth]{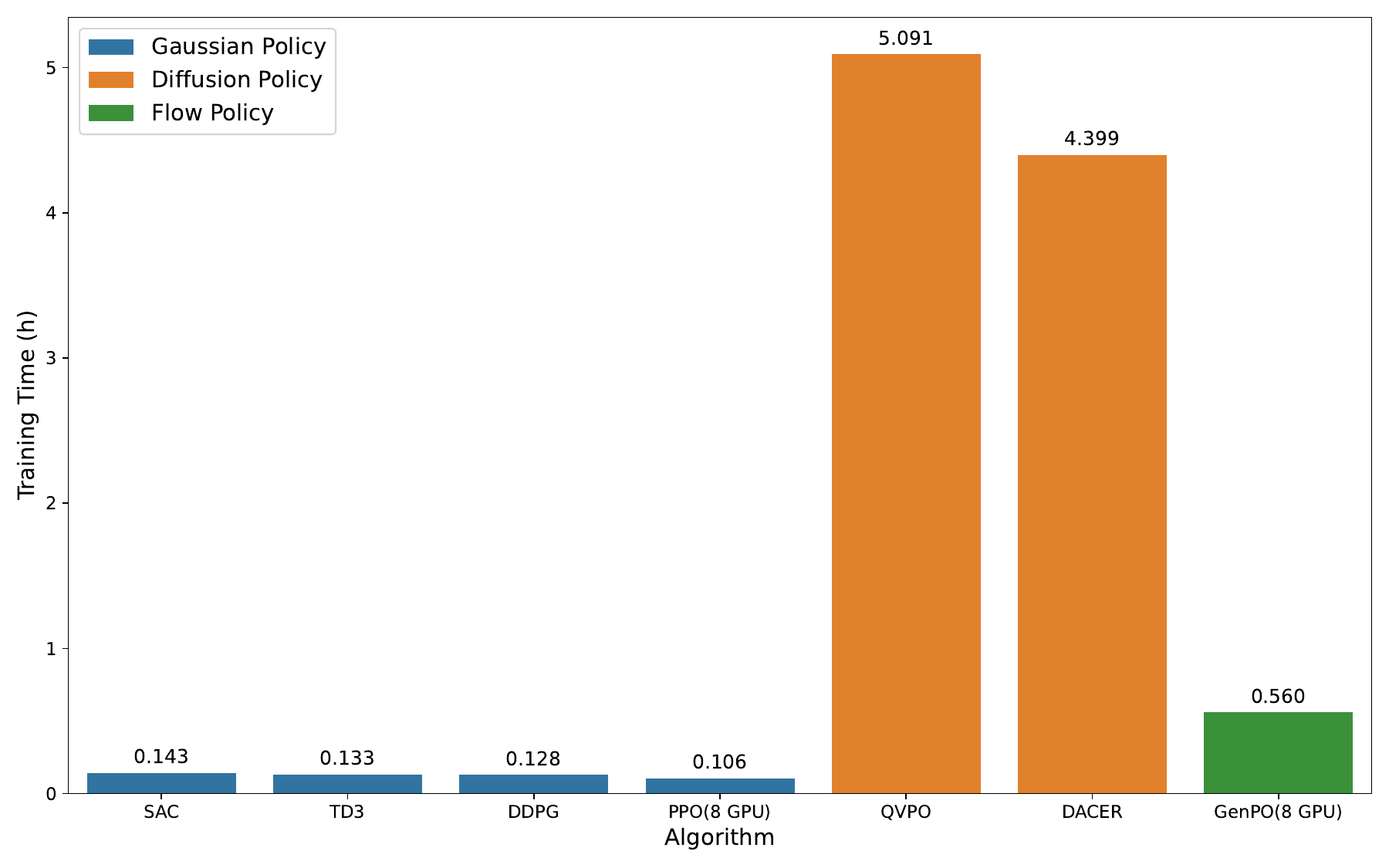}
    \caption{Training time comparison on Isaac-Ant-v0 environment.}
    \label{fig:train}
\end{figure}

\begin{figure}[ht]
    \centering
    \includegraphics[width=1.0\linewidth]{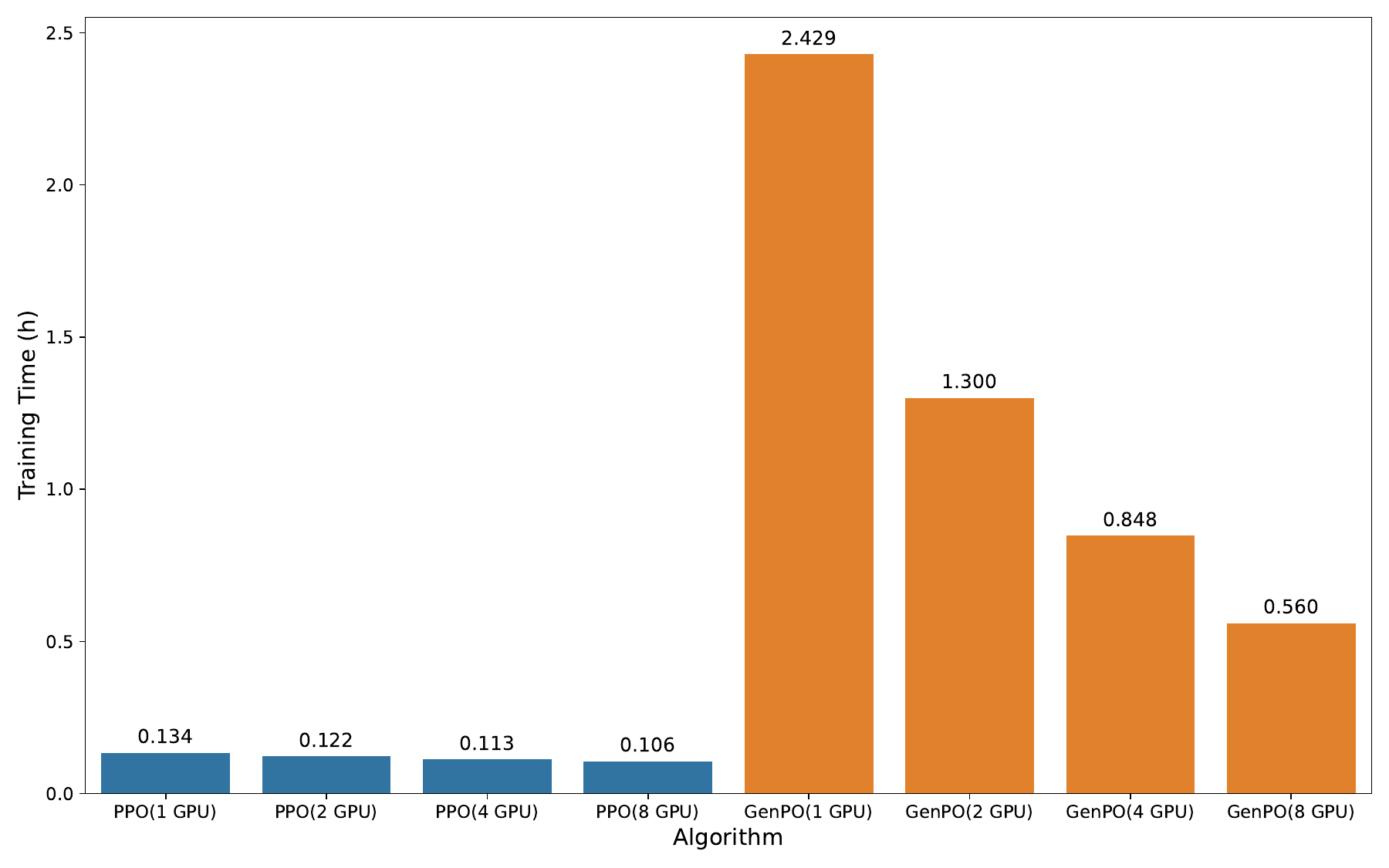}
    \caption{Comparison of training time between PPO and GenPO on the Isaac-And-v0 environment for scaling training on multiple GPUs.}
    \label{fig:num_envs}
\end{figure}

To validate the computational efficiency, we also conduct comparison experiments on the proposed GenPO with 6 other baselines, which respectively evaluate the inference, training and parallel training time on single/multiple GPUs, as shown in Figure~\ref{fig:inference}, Figure~\ref{fig:train} and Figure~\ref{fig:num_envs}. It can be observed that, while the  GenPO's inference time is slightly longer than that of Gaussian policies due to multiple denoising iterations of diffusion, it remains significantly more efficient than previous diffusion-based RL algorithms such as QVPO and DACER. Besides, to our knowledge, the inference time of $2.577$ms is more than sufficient for the real-time decision in robot control. Furthermore, in terms of training time, although the performance of GenPO on a single GPU is less than ideal (yet still superior to previous diffusion-based RL methods), it can effectively leverage multi-GPU parallelism to significantly accelerate training. In contrast, the traditional RL algorithm PPO cannot obviously benefit from that.

\newpage

\section{Proof for Lemma 1}

\textbf{Lemma 1.}
% \label{lemma 1}
\textit{\textbf{(Change of Variables ~\cite{blitzstein2019introduction})} Let $f: \mathbb{R}^n\rightarrow \mathbb{R}^n$ is an invertible and smooth mapping. If we have the random variable $X\sim q(x)$ and the random variable $Y=f(X)$ transformed by function $f$, the distribution $p(y)$ of $Y$ is
\begin{equation}
    p(y)=q(x)\left|\det\frac{\partial f}{\partial x}\right|^{-1}.
\end{equation}
}

\begin{proof}
Here, we take the one-dimensional random variables $X, Y$ as an example. For a general n-dimensional case, we can obtain the same result by applying multivariable calculus.

Firstly, we can divide $f$ into two categories: (1) $f$ is strictly increasing, (2) $f$ is strictly decreasing, since $f$ is an invertible and smooth mapping.

\textbf{Strictly increasing.} For strictly increasing $f$, the CDF of  $Y$  is
\[
P(y)=P(Y \leq y)=P(f(X) \leq y)=P\left(X \leq f^{-1}(y)\right)=Q\left(f^{-1}(y)\right)=Q(x).
\]
Then, according to $p(x)=\lim\limits_{\epsilon\to 0}\frac{P(x+\epsilon)-P(x)}{\epsilon}$ and the chain rule, the PDF of the random variable $Y$ is
\[
p(y)=q(x)\frac{dx}{dy}=q(x)\left(\det\frac{\partial f}{\partial x}\right)^{-1}.
\]
\textbf{Strictly decreasing.} The proof for  $f$  strictly decreasing is analogous. In that case, the PDF of $Y$ is  $p(y)=-q(x) \frac{dx}{dy}$ since $\frac{dx}{dy}<0$. Hence, we need to add an absolute sign for the final formula, i.e., $    p(y)=q(x)\left|\det\frac{\partial f}{\partial x}\right|^{-1}$.
\end{proof}
%%%%%%%%%%%%%%%%%%%%%%%%%%%%%%%%%%%%%%%%%%%%%%%%%%%%%%%%%%%%

\clearpage
\section*{NeurIPS Paper Checklist}

\begin{enumerate}

\item {\bf Claims}
    \item[] Question: Do the main claims made in the abstract and introduction accurately reflect the paper's contributions and scope?
    \item[] Answer: \answerYes{} % Replace by \answerYes{}, \answerNo{}, or \answerNA{}.
    \item[] Justification: As stated in the abstract and introduction, this paper bridges the gap between on-policy reinforcement learning and generative diffusion models, makes it possible to calculate the log-likelihood in diffusion policies, and finally enables large-scale parallel simulation environments to leverage the powerful multimodal capabilities of diffusion models. They can accurately reflect the contributions and scope of this paper.

\item {\bf Limitations}
    \item[] Question: Does the paper discuss the limitations of the work performed by the authors?
    \item[] Answer: \answerYes{} % Replace by \answerYes{}, \answerNo{}, or \answerNA{}.
    \item[] Justification: The analysis of limitations is provided in Section~\ref{sect:lim}.

\item {\bf Theory assumptions and proofs}
    \item[] Question: For each theoretical result, does the paper provide the full set of assumptions and a complete (and correct) proof?
    \item[] Answer: \answerYes{} % Replace by \answerYes{}, \answerNo{}, or \answerNA{}.
    \item[] Justification: The proofs are provided in the supplementary materials.

    \item {\bf Experimental result reproducibility}
    \item[] Question: Does the paper fully disclose all the information needed to reproduce the main experimental results of the paper to the extent that it affects the main claims and/or conclusions of the paper (regardless of whether the code and data are provided or not)?
    \item[] Answer: \answerYes{} % Replace by \answerYes{}, \answerNo{}, or \answerNA{}.
    \item[] Justification: These details are provided in the supplementary materials.

\item {\bf Open access to data and code}
    \item[] Question: Does the paper provide open access to the data and code, with sufficient instructions to faithfully reproduce the main experimental results, as described in supplemental material?
    \item[] Answer: \answerYes{} % Replace by \answerYes{}, \answerNo{}, or \answerNA{}.
    \item[] Justification: We will release the code once the paper is accepted.

\item {\bf Experimental setting/details}
    \item[] Question: Does the paper specify all the training and test details (e.g., data splits, hyperparameters, how they were chosen, type of optimizer, etc.) necessary to understand the results?
    \item[] Answer: \answerYes{} % Replace by \answerYes{}, \answerNo{}, or \answerNA{}.
    \item[] Justification: These details are provided in Section~\ref{sect: experiment} and supplementary materials.

\item {\bf Experiment statistical significance}
    \item[] Question: Does the paper report error bars suitably and correctly defined or other appropriate information about the statistical significance of the experiments?
    \item[] Answer: \answerYes{} % Replace by \answerYes{}, \answerNo{}, or \answerNA{}.
    \item[] Justification: The standard deviation in our experiment results (Figure~\ref{fig:comparison}, Figure~\ref{fig:ablation} and Table~\ref{tab:result}) shows the statistical significance.

\item {\bf Experiments compute resources}
    \item[] Question: For each experiment, does the paper provide sufficient information on the computer resources (type of compute workers, memory, time of execution) needed to reproduce the experiments?
    \item[] Answer: \answerYes{} % Replace by \answerYes{}, \answerNo{}, or \answerNA{}.
    \item[] Justification: The information on the computer resources is provided in the supplementary materials.
    
\item {\bf Code of ethics}
    \item[] Question: Does the research conducted in the paper conform, in every respect, with the NeurIPS Code of Ethics \url{https://neurips.cc/public/EthicsGuidelines}?
    \item[] Answer: \answerYes{} % Replace by \answerYes{}, \answerNo{}, or \answerNA{}.
    \item[] Justification: Our research conformed with the NeurIPS Code of Ethics.

\item {\bf Broader impacts}
    \item[] Question: Does the paper discuss both potential positive societal impacts and negative societal impacts of the work performed?
    \item[] Answer: \answerNA{} % Replace by \answerYes{}, \answerNo{}, or \answerNA{}.
    \item[] Justification: Our paper primarily focuses on theoretical research in how to employ the diffusion model in on-policy reinforcement learning, with no consideration of societal impacts.
    
\item {\bf Safeguards}
    \item[] Question: Does the paper describe safeguards that have been put in place for responsible release of data or models that have a high risk for misuse (e.g., pretrained language models, image generators, or scraped datasets)?
    \item[] Answer: \answerNA{} % Replace by \answerYes{}, \answerNo{}, or \answerNA{}.
    \item[] Justification: Our experiment of this paper was conducted on 8 robot control tasks in IsaacLab, which poses no such risks.

\item {\bf Licenses for existing assets}
    \item[] Question: Are the creators or original owners of assets (e.g., code, data, models), used in the paper, properly credited and are the license and terms of use explicitly mentioned and properly respected?
    \item[] Answer: \answerYes{} % Replace by \answerYes{}, \answerNo{}, or \answerNA{}.
    \item[] Justification: The implementation of SAC, TD3, DDPG, PPO, DACER, and QVPO, and IsaacLab simulator are cited properly in the supplementary materials.

\item {\bf New assets}
    \item[] Question: Are new assets introduced in the paper well documented and is the documentation provided alongside the assets?
    \item[] Answer: \answerNA{} % Replace by \answerYes{}, \answerNo{}, or \answerNA{}.
    \item[] Justification: The paper does not introduce new assets.

\item {\bf Crowdsourcing and research with human subjects}
    \item[] Question: For crowdsourcing experiments and research with human subjects, does the paper include the full text of instructions given to participants and screenshots, if applicable, as well as details about compensation (if any)? 
    \item[] Answer: \answerNA{} % Replace by \answerYes{}, \answerNo{}, or \answerNA{}.
    \item[] Justification: The paper does not involve crowdsourcing nor research with human subjects.

\item {\bf Institutional review board (IRB) approvals or equivalent for research with human subjects}
    \item[] Question: Does the paper describe potential risks incurred by study participants, whether such risks were disclosed to the subjects, and whether Institutional Review Board (IRB) approvals (or an equivalent approval/review based on the requirements of your country or institution) were obtained?
    \item[] Answer: \answerNA{} % Replace by \answerYes{}, \answerNo{}, or \answerNA{}.
    \item[] Justification: The paper does not involve crowdsourcing nor research with human subjects.

\item {\bf Declaration of LLM usage}
    \item[] Question: Does the paper describe the usage of LLMs if it is an important, original, or non-standard component of the core methods in this research? Note that if the LLM is used only for writing, editing, or formatting purposes and does not impact the core methodology, scientific rigorousness, or originality of the research, declaration is not required.
    %this research? 
    \item[] Answer: \answerNA{} % Replace by \answerYes{}, \answerNo{}, or \answerNA{}.
    \item[] Justification: We merely use the LLM for editing the paper.

\end{enumerate}

\end{document}